\documentclass[11pt]{article}

\usepackage[final]{acl}

\usepackage{enumitem}

\usepackage{times}
\usepackage{latexsym}
\usepackage{amsmath}

\usepackage{multirow}
\usepackage{colortbl}
\usepackage{xcolor}
\definecolor{bettergreen}{rgb}{0.85, 0.95, 0.85}
\definecolor{worsecolor}{rgb}{1.0, 0.85, 0.85}
\definecolor{graycolor}{rgb}{0.96, 0.96, 0.96}
\definecolor{mathcolor}{rgb}{0.85, 0.91, 0.98}

\usepackage[T1]{fontenc}
\usepackage[utf8]{inputenc}

\usepackage{microtype}

\usepackage{inconsolata}

\usepackage{graphicx}

\title{From Implicit to Explicit: Token-Efficient Logical Supervision for Mathematical Reasoning in LLMs}

\author{
  \textbf{Shaojie Wang\textsuperscript{1}} \quad
  \textbf{Liang Zhang\textsuperscript{1}}\thanks{\ Corresponding author.} \\
  \\
  \texttt{\{shaojiewang, liangzhang\}@hkustgz.edu.cn} \\
  \\
  \textsuperscript{1} Hong Kong University of Science and Technology (Guangzhou) \\
}

\begin{document}
\maketitle
\begin{abstract}
Recent studies reveal that large language models (LLMs) exhibit limited logical reasoning abilities in mathematical problem-solving, instead often relying on pattern-matching and memorization. We systematically analyze this limitation, focusing on logical relationship understanding, which is a core capability underlying genuine logical reasoning, and reveal that errors related to this capability account for over 90\% of incorrect predictions, with Chain-of-Thought Supervised Fine-Tuning (CoT-SFT) failing to substantially reduce these errors. To address this bottleneck, we propose \textbf{F}irst-\textbf{S}tep \textbf{L}ogical \textbf{R}easoning (\textbf{FSLR}), a lightweight training framework targeting logical relationship understanding.  Our key insight is that the first planning step-identifying which variables to use and which operation to apply without performing any calculation-encourages the model to derive logical relationships directly from the problem statement. By training models on this isolated step, FSLR provides explicit supervision for logical relationship understanding, unlike CoT-SFT which implicitly embeds such relationships within complete solution trajectories. Extensive experiments across multiple models and datasets demonstrate that FSLR consistently outperforms CoT-SFT under both in-distribution and out-of-distribution settings, with average improvements of 3.2\% and 4.6\%, respectively. Moreover, FSLR achieves 4-6× faster training and reduces training token consumption by over 80\%.

% while achieving 4-6× faster training and reducing training tokens by over 80\%.

\end{abstract}

\section{Introduction}
Mathematical reasoning, as a crucial cognitive skill that supports problem-solving in numerous scientific and practical applications, has attracted particular attention in LLM research~\cite{llm4mathreasoning,llm_mathematical_survey,llm_mathematical_optimization_survey}. Although LLMs have achieved near-human accuracy on mathematical benchmarks such as GSM8K~\cite{gsm8k}, recent studies~\cite{gsm_symbolic,math_perturb,gsm_plus} reveal that LLMs exhibit limited logical reasoning abilities—the capacity to understand "novel" problems and derive appropriate solution steps, rather than relying on pattern-matching and memorization.

\noindent As a fundamental capability for mathematical problem-solving, genuine logical reasoning enables models to understand the underlying rationale of problems and transfer knowledge to new situations~\cite{logical_reasoning}. 
% While prior work has primarily examined reasoning capabilities across pre-trained LLMs, their fine-tuned counterparts have received limited attention. 
Recently, Supervised Fine-Tuning (SFT) with chain-of-thought (CoT) supervision has received increasing attention and has been widely adopted to improve reasoning abilities of LLMs by training models on complete solution trajectories~\cite{reasoning_survey}. However, it remains unclear whether CoT-SFT can elicit genuine logical reasoning.

\begin{figure}[h]
  \includegraphics[width=\columnwidth]{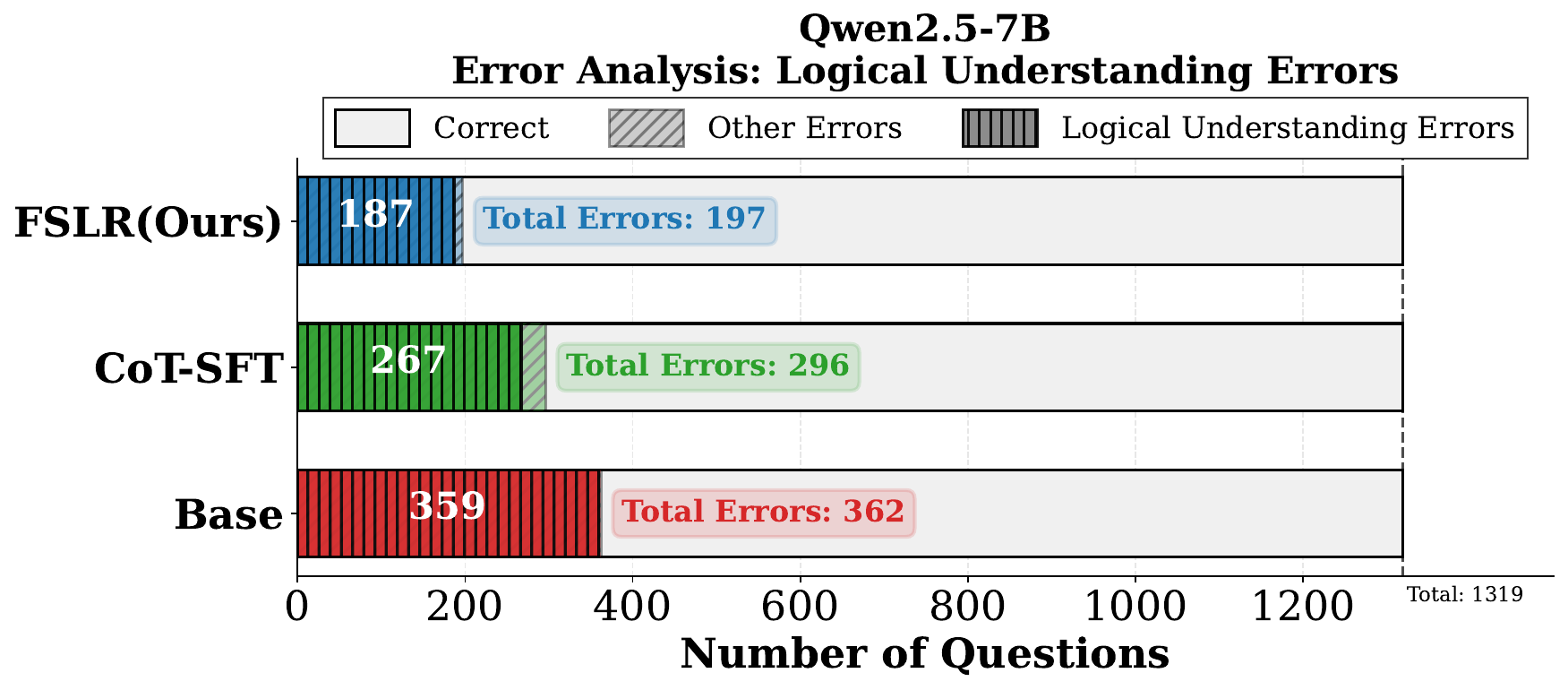}
  \caption{Error analysis on Qwen2.5-7B comparing Base, CoT-SFT, and FSLR(Ours) models. Each bar shows the breakdown of correct predictions, logical relationship understanding errors, and other errors.}
  \label{fig:failure_portion}
\end{figure}

\noindent To investigate this question, we conduct a systematic analysis comparing pretrained LLMs with their fine-tuned counterparts on mathematical reasoning tasks. Following the above discussion, genuine logical reasoning requires understanding a given problem before deriving solution steps. For mathematical problems, such understanding involves grasping how variables in the problem relate to each other 
and which operations connect them—what we term logical relationships. 
% \textit{logical relationships between variables} (hereafter, \textit{logical relationships}). 
We refer to the ability to grasp these relationships as logical relationship understanding, and adopt it as our diagnostic metric for genuine logical reasoning. Specifically, we use GPT-4o~\cite{gpt4o} to analyze all incorrect predictions and identify whether failures arise from misunderstanding these logical relationships(Appendix~\ref{sec:error-analysis-prompt}). As shown in Figure~\ref{fig:failure_portion}, using Qwen2.5-7B as an example, we find that logical relationship understanding errors account for over 90\% of incorrect predictions—a pattern that holds consistently across multiple LLMs (Appendix~\ref{error}).
% ; error classification prompt in Appendix~\ref{sec:error-analysis-prompt}).
Since CoT-SFT fails to substantially reduce these errors, they become the primary bottleneck limiting further improvements in overall accuracy.

\noindent These findings raise a natural research question: how can we more directly improve LLMs' understanding of logical relationships, thereby enhancing their overall performance? To answer this question, we examine why CoT-SFT falls short. CoT-SFT trains models to imitate complete solution trajectories, while logical relationships are implicitly embedded in these trajectories, never directly targeted by the training objective. We hypothesize that this implicit supervision leads to insufficient learning of logical relationships and consequently results in the phenomenon illustrated in Figure~\ref{fig:failure_portion}.

\noindent To address this limitation, we propose \textbf{F}irst-\textbf{S}tep \textbf{L}ogical \textbf{R}easoning (\textbf{FSLR}), a lightweight training framework that provides explicit supervision for enhancing logical relationship understanding. 
% Unlike CoT-SFT, FSLR isolates the initial planning decision as a standalone training task. 
Specifically, we design a first-planning-step prompt schema that asks models to identify only what needs to be calculated first—which variables to use and which operation to apply, without solving the full problem. Then, FSLR isolates this first planning step as a standalone training task, thereby providing explicit supervision for logical relationship understanding.

\noindent This design offers several key advantages: (1) \textbf{More focused supervision}: FSLR provides a more direct training signal for logical relationship understanding by isolating the first planning step explicitly,
% from computational execution, 
alleviating the core limitation we identified earlier; (2) \textbf{Reduced training cost}: predicting only the initial planning step requires significantly fewer tokens than complete trajectories; (3) \textbf{Improved generalization}: since logical relationship understanding forms the foundation of the entire reasoning chain, strengthening this ability is expected to benefit the entire problem-solving process. As shown in Figure~\ref{fig:failure_portion}, FSLR achieves a substantial reduction in both logical relationship understanding errors and overall errors compared to CoT-SFT. Our main contributions are as follows:
\begin{itemize}
\item We systematically analyze the reasoning capabilities of LLMs and identify logical relationship understanding as a critical bottleneck: such errors account for over 90\% of failures, and CoT-SFT fails to substantially mitigate this issue.

% We systematically analyze the reasoning capabilities of LLMs, revealing that logical relationship understanding errors account for over 90\% of failures and CoT-SFT fails to substantially reduce this proportion.

\item We propose FSLR, a lightweight training framework that provides a more focused training task for logical relationship understanding by training models to identify the initial planning step obtained under a specially designed prompting scheme.
% what to compute first.

\item Extensive experiments across multiple models and datasets demonstrate that FSLR substantially outperforms CoT-SFT in both in-distribution and out-of-distribution settings, while requiring significantly fewer training tokens. 
\end{itemize}

% Consider what happens when a model encounters a novel math problem; before any computation, it must determine where to begin: which quantities to combine, which relationships to exploit, or which operation to apply. 
% % we need to 

% This initial decision requires understanding how variables relate to each other, which is the logical relationship we aims to train. 

% Specifically, at this step, no intermediate computational results are available; the model cannot rely on prior calculated values as cues and is instead encouraged to identify logical relationship largely from the problem statement. By training model to produce this initial planning step without executing the subsequent calculations, we can isolate logical relationship understanding as the sole training objective. This insight motivates our approach: targeting this critical initial reasoning phase as an explicit training objective.

\section{Related Work}
\subsection{Mathematical Reasoning in LLMs}
Mathematical reasoning has emerged as a critical testbed for evaluating whether large language models possess genuine reasoning abilities. Driven by this goal, the community has introduced diverse benchmarks~\cite{gsm8k,svamp,asdiv,mawps,tabmwp,gsm_hard}, on which recent LLMs have achieved remarkable success. However, recent studies question whether these successes reflect genuine logical reasoning. GSM-Symbolic~\cite{gsm_symbolic} and GSM-Plus~\cite{gsm_plus} reveal that LLMs exhibit noticeable performance variance across problem instantiations and are far from robust. Other work shows that LLMs are easily distracted by irrelevant information~\cite{gsm_ic} and blindly apply learned skills without assessing their applicability to modified contexts~\cite{math_perturb}. While these works clearly diagnose that current LLMs often rely on pattern matching rather than true reasoning, they do not propose concrete training strategies to improve this capability.

% logical relationship understanding.

\subsection{Chain-of-Thought Fine-Tuning}
Chain-of-Thought Supervised Fine-Tuning (CoT-SFT) has emerged as a widely adopted approach for enhancing LLM reasoning by exposing models to step-by-step solution trajectories~\cite{self_explore_sft,teaching_to_small_llm,learning_composable_cot,self_train_meets_consistency}. However, recent studies reveal several limitations: CoT-SFT decreases the faithfulness of reasoning~\cite{impact_cot_sft}, leads models to memorize task-specific templates rather than acquiring transferable abilities~\cite{sft_memorize}, and introduces spurious features that cause hallucinations~\cite{llm_mimic_human}. While these studies reveal important limitations of CoT-SFT, none directly targets the logical relationship understanding bottleneck we identified.

\section{Methodology}

\subsection{Task Formulation}

We formalize CoT-SFT and FSLR to clarify how they differ in supervising logical relationship understanding.

\noindent \textbf{CoT-SFT.} Given a mathematical problem $p$ and its complete solution trajectory $s = (s_1, s_2, ..., s_n)$, CoT-SFT trains models to generate the full reasoning chain by optimizing:
$$\mathcal{L}_{\text{CoT}} = -\sum_{i=1}^{n} \log P(s_i | p, s_1, ..., s_{i-1}; \theta)$$
While logical relationships are embedded within these steps, they are never explicitly targeted by the training objective. The supervision signal is distributed across the entire trajectory, diluting the focus on logical relationship understanding, we refer to this paradigm as \textit{implicit supervision}.

\noindent \textbf{FSLR.} Given the same problem $p$, FSLR trains models to generate only the first planning step $f_1$ by optimizing:
$$\mathcal{L}_{\text{FSLR}} = -\log P(f_1 | p; \theta)$$
where $f_1$ identifies which variables to use and which operation to apply, corresponding to the first planning step detailed in the next section. By isolating logical relationship understanding as an explicit training objective, our formulation directly targets this capability. We refer to this paradigm as \textit{explicit supervision}.

% Unlike complete solution trajectories where logical relationships are interspersed among multiple reasoning steps, 

\begin{figure}[t]
  \centering
  \includegraphics[width=0.95\columnwidth]{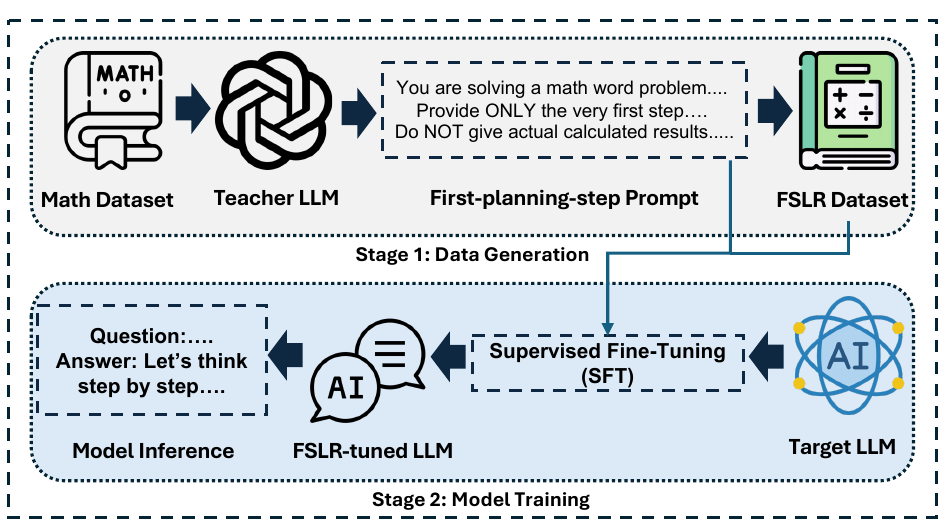}
   \caption{Overview of the FSLR framework, consisting of two stages: data generation and model training. The framework leverages a teacher LLM to generate first planning step guidance, which is then used to fine-tune the target LLM via supervised fine-tuning.}
  \label{fig:framework}
\end{figure}

\subsection{FSLR Framework}

Based on the above analysis, we now present the FSLR framework in detail. Specifically, FSLR trains on the first planning step—identifying relevant variables and operations without performing calculations—rather than the first calculation step. Corsidering logical relationship understanding forms the foundation of the entire reasoning chain, once the model correctly identifies which variables to use and which operation to apply, subsequent arithmetic steps follow naturally. As illustrated in Figure~\ref{fig:framework}, FSLR consists of two stages: \textbf{(1)} constructing a training dataset $\mathcal{D}_{\text{FSLR}}$ that isolates logical relationship understanding, and \textbf{(2)} fine-tuning target models to generate $f_1$, providing explicit supervision for this core capability.

\subsubsection{Training Dataset Construction}
\label{data:prompt}
We construct training dataset $\mathcal{D}_{\text{FSLR}}$ by prompting a teacher model to generate first planning step for mathematical problems.

\noindent \textbf{First-planning-step Prompt Design.} Given a problem $p$, we use the following prompt template:
\begin{quote}
\small\ttfamily
You are solving a math word problem step by step. Your task is to provide ONLY the very first step - stop immediately after identifying what to calculate first.
Rules:
\begin{itemize}
\item Provide ONLY the first calculation or identification
\item Do NOT solve the entire problem
\item Do NOT provide multiple steps
\item Do NOT give the actual calculated result, just identify what needs to be calculated
\end{itemize}
Problem: [problem text]
First Step Only:
\end{quote}

\noindent \textbf{Design Rationale.} The prompt instructs the model to identify only which variables to use and which operation to apply, ensuring that $f_1$ captures purely the logical reasoning decision. This aligns with our goal of providing explicit supervision for logical relationship understanding.

\noindent \textbf{Data Generation.} For each problem $p_i$, we query the teacher model to generate $f_1^{(i)}$ using the designed prompt and construct the training dataset $\mathcal{D}_{\text{FSLR}} = \{(p_i, f_1^{(i)})\}_{i=1}^{N}$. We provide an example illustrating the training data format in Appendix~\ref{appendix:training-example}.
% We filter out responses containing complete solutions. 

\subsubsection{Model Training}
Given $\mathcal{D}_{\text{FSLR}}$, the target LLM is fine-tuned to maximize the likelihood of $f_1^{(i)}$ conditioned on the input $\text{prompt}(p_i)$ as follows:
$$\mathcal{L}_{\text{FSLR}} = -\frac{1}{N}\sum_{i=1}^{N} \log P(f_1^{(i)} | \text{prompt}(p_i); \theta)$$
where $\text{prompt}(p_i)$ denotes $p_i$ formatted with our prompt template, and $f_1^{(i)}$ is the first planning step for problem $p_i$. Through this training process, the model learns to explicitly identify logical relationships between variables. Notably, since FSLR trains only on the first planning step $f_i$, it requires significantly fewer tokens than CoT-SFT, substantially reducing training costs.

\subsubsection{Model Inference}
Since FSLR strengthens logical relationship understanding: the foundation of the entire reasoning chain, the improved capability is expected to benefit the full problem-solving process. At inference time, the FSLR-trained model generates complete solutions through standard autoregressive decoding, with no specialized prompting or additional modules required. 

\begin{table*}[h]
  \centering
  \normalsize
  \setlength{\tabcolsep}{4pt} 
  % Add these to preamble:
  % \usepackage{colortbl}
  % \usepackage{xcolor}
  % \definecolor{bettergreen}{rgb}{0.85, 0.95, 0.85}
  % \definecolor{worsecolor}{rgb}{1.0, 0.85, 0.85}
  \resizebox{0.95\textwidth}{!}{%
  \begin{tabular}{l|l|cc|cc|cc|c}
    \hline
    \multicolumn{2}{c|}{\cellcolor{graycolor}\textbf{In-Distribution Results}} & \multicolumn{2}{c|}{\textit{\cellcolor{graycolor}\textbf{LLaMA3.1-8B}}} & \multicolumn{2}{c|}{\cellcolor{graycolor}\textit{\textbf{Qwen2.5-7B}}} & \multicolumn{2}{c|}{\cellcolor{graycolor}\textit{\textbf{Qwen3-4B}}} & 
    \multirow{2}{*}{\textbf{Average}} \\
    \cline{1-8}
   \textbf{Data Source} & \textbf{Method} & \textbf{GSM8K} & \textbf{SVAMP} &\textbf{GSM8K} & \textbf{SVAMP} & \textbf{GSM8K} &\textbf{SVAMP} & \\
    \hline
    \hline
    \textbf{DeepSeek-Math} & \cellcolor{graycolor}Zero-shot & \cellcolor{mathcolor}78.70 & \cellcolor{mathcolor}82.20 & \cellcolor{mathcolor}78.70 & \cellcolor{mathcolor}82.20 & \cellcolor{mathcolor}78.70 & \cellcolor{mathcolor}82.20 & \cellcolor{mathcolor}80.45 \\
    \textbf{Qwen2.5-Math}& Zero-shot & \cellcolor{mathcolor}79.20 & \cellcolor{mathcolor}85.50 & \cellcolor{mathcolor}79.20 & \cellcolor{mathcolor}85.50 & \cellcolor{mathcolor}79.20 & \cellcolor{mathcolor}85.50 & \cellcolor{mathcolor}82.35
    \\
    \hline
    \multirow{4}{*}{\textbf{Base LLM}} 
    & \cellcolor{graycolor}Zero-shot & \cellcolor{graycolor}62.90 & \cellcolor{graycolor}67.60 & \cellcolor{graycolor}72.60 & \cellcolor{graycolor}83.00 & \cellcolor{graycolor}84.70 & \cellcolor{graycolor}85.00 & \cellcolor{graycolor}75.97 \\
    & Few-shot & 77.50 & 84.00 & 90.10 & 92.20 & 84.80 & 91.60 & 86.70 \\
    & Zero-LP & \cellcolor{graycolor}70.00 & \cellcolor{graycolor}72.30 & \cellcolor{graycolor}84.50 &\cellcolor{graycolor} 88.50 & \cellcolor{graycolor}86.80 & \cellcolor{graycolor}92.60 & \cellcolor{graycolor}82.45 \\
    & Few-LP  & 67.70 & 71.80 & 86.20 & 87.30 & 91.30 & 91.70 & 82.67 \\ 
    \hline

    \multirow{2}{*}{\textbf{LLaMA}} 
    & CoT-SFT & \cellcolor{bettergreen}77.90 & \cellcolor{bettergreen}79.30 & \cellcolor{bettergreen}77.60 & \cellcolor{bettergreen}88.10 & \cellcolor{bettergreen}85.70 & \cellcolor{bettergreen}85.10 & \cellcolor{bettergreen}82.28 \\
    & \cellcolor{graycolor}FSLR & \cellcolor{bettergreen}\textbf{83.10} & \cellcolor{bettergreen}\textbf{84.80} & \cellcolor{bettergreen}\textbf{85.10} & \cellcolor{bettergreen}\textbf{91.30} & \cellcolor{bettergreen}\textbf{87.10} & \cellcolor{bettergreen}\textbf{91.10} & \cellcolor{bettergreen}\textbf{87.08} \\ 
    \hline
    \multirow{2}{*}{\textbf{Qwen}} 
    & CoT-SFT & \cellcolor{worsecolor}\textbf{85.70} & \cellcolor{bettergreen}84.00 & \cellcolor{bettergreen}82.60 & \cellcolor{bettergreen}89.30 & \cellcolor{bettergreen}91.30 & \cellcolor{worsecolor}\textbf{93.90} & \cellcolor{bettergreen}87.80 \\
    & \cellcolor{graycolor}FSLR & \cellcolor{worsecolor}85.30 & \cellcolor{bettergreen}\textbf{85.90} & \cellcolor{bettergreen}\textbf{86.40} & \cellcolor{bettergreen}\textbf{92.70} & \cellcolor{bettergreen}\textbf{91.80} & \cellcolor{worsecolor}91.30 & \cellcolor{bettergreen}\textbf{88.90} \\ 
    \hline
    \multirow{2}{*}{\textbf{Self}} 
    & CoT-SFT & \cellcolor{bettergreen}74.10 & \cellcolor{bettergreen}78.80 & \cellcolor{bettergreen}84.50 & \cellcolor{bettergreen}90.80 & \cellcolor{bettergreen}89.80 & \cellcolor{bettergreen}90.10 & \cellcolor{bettergreen}84.68 \\
    & \cellcolor{graycolor}FSLR & \cellcolor{bettergreen}\textbf{80.40} & \cellcolor{bettergreen}\textbf{82.50} & \cellcolor{bettergreen}\textbf{88.90} & \cellcolor{bettergreen}\textbf{93.00} & \cellcolor{bettergreen}\textbf{92.10} & \cellcolor{bettergreen}\textbf{94.00} & \cellcolor{bettergreen}\textbf{88.48} \\
    \hline
  \end{tabular}
  }
  \caption{In-distribution evaluation on GSM8K and SVAMP. Models are trained on GSM8K and SVAMP using data generated by different teachers (LLaMA-3.1-70B, Qwen2.5-72B, or self-generated). All trained models are evaluated under zero-shot setting. Zero-LP and Few-LP denote zero-shot and few-shot logical planning prompting strategies that explicitly instruct models to identify logical relationships before solving, without fine-tuning. Math-specialized models (DeepSeek-Math-7B and Qwen2.5-Math-7B) are evaluated zero-shot as reference baselines.  \colorbox{bettergreen}{Green cells} indicate FSLR outperforms CoT-SFT. \colorbox{worsecolor}{Red cells} indicate FSLR underperforms CoT-SFT.}
  \label{tab:in-domain}
\end{table*}

\section{Experiments}
In this section, we present comprehensive experiments to evaluate the effectiveness of FSLR training. We first describe the experimental setup  (Section~\ref{sec:exp_setup}), then present our experiments results (Section~\ref{sec:exp_results}).

% and provide case studies illustrating the improvements .

\subsection{Experimental Setup}
\label{sec:exp_setup}

\noindent\textbf{Datasets:} We evaluate FSLR on multiple mathematical reasoning benchmarks to assess both in-distribution and out-of-distribution performance. For \textbf{in-distribution evaluation}, we use GSM8K~\cite{gsm8k} and SVAMP~\cite{svamp}. For \textbf{out-of-distribution evaluation}, we test on four additional datasets: ASDiv~\cite{asdiv}, MAWPS~\cite{mawps} , TabMWP~\cite{tabmwp}, and GSM-Hard~\cite{gsm_hard}.

\noindent\textbf{Models.} We conduct experiments across multiple model families to ensure generalizability. For \textbf{training data generation}, we use three teacher models: LLaMA-3.1-70B-Instruct~\cite{llama3.1}, Qwen2.5-72B-Instruct~\cite{qwen2}, and the finetuned target model itself (self-generated). For \textbf{target models}, we fine-tune three instruction-tuned checkpoints: LLaMA-3.1-8B-Instruct~\cite{llama3.1}, Qwen2.5-7B-Instruct~\cite{qwen2}, and Qwen3-4B-Instruct~\cite{qwen3}.
\noindent\textbf{Baselines.} We compare FSLR against the following baselines:
\begin{itemize}
\item \textbf{Base Model}: The original instruction-tuned model without additional mathematical reasoning training.
\item \textbf{Math-Specialized Model}: Representative models fine-tuned specifically for mathematical reasoning at comparable scale, including DeepSeek-Math-7B-Instruct ~\cite{deepseek_math} and Qwen2.5-Math-7B~\cite{qwen2.5_math}.
\item \textbf{CoT-SFT Model}: Models fine-tuned on complete chain-of-thought solution trajectories, representing the current mainstream approach.
\item \textbf{Zero-LP}: A zero-shot prompting strategy that instructs the model to first identify relevant variables and operations before solving, without fine-tuning.
\item \textbf{Few-LP}: A few-shot prompting strategy that provides demonstrations of explicit logical relationship identification before solving, without fine-tuning.
\end{itemize}

\noindent\textbf{Training Details.} For both CoT-SFT and FSLR, we use the complete training sets of GSM8K and SVAMP as our base datasets. For CoT-SFT data generation, we use two sources: (1) \textbf{Teacher LLM}: a larger model is prompted to generate complete step-by-step solution trajectories for each training problem; (2) \textbf{Self-generated}: the target model itself generates its own CoT solutions using the same prompt template. In both cases, we use greedy decoding (temperature=0) via vLLM for efficient batch inference, and retain only solutions yielding the correct final answer. For FSLR, we generate first planning steps using the prompt template described in Section~\ref{data:prompt}.

\noindent\textbf{Evaluation Protocol.} We evaluate all models using greedy decoding with the standard ``Let's think step by step'' prompting strategy and report accuracy as the primary metric.

\begin{table*}[!h]
  \centering
  \normalsize
  \setlength{\tabcolsep}{6pt}
  % Add these to preamble:
  % \usepackage{colortbl}
  % \usepackage{xcolor}
  % \definecolor{bettergreen}{rgb}{0.85, 0.95, 0.85}
  % \definecolor{worsecolor}{rgb}{1.0, 0.85, 0.85}
  \resizebox{0.90\textwidth}{!}{%
  \begin{tabular}{l|l|ccccc|c}
    \hline
    \multicolumn{2}{c|}{\cellcolor{graycolor}\textbf{Out-of-Distribution Results}} & \multicolumn{5}{c|}{\cellcolor{graycolor}\textbf{Models Trained on GSM8K}} & \multirow{2}{*}{\textbf{Average}} \\
    \cline{1-7}
    \textbf{Data Source} & \textbf{Method} & \textbf{AsDIv} & \textbf{SVAMP} & \textbf{MAWPS} & \textbf{TabMWP} & \textbf{GSM-Hard} & \\
    \hline
    \hline
    \multicolumn{8}{c}{\textit{\textbf{Math-Specialized Models}}} \\
    \hline
    \cellcolor{graycolor}\textbf{DeepSeek-Math} & \cellcolor{graycolor}Zero-shot & \cellcolor{mathcolor}85.00 &\cellcolor{mathcolor}82.20	&\cellcolor{mathcolor}92.50	&\cellcolor{mathcolor}69.90	&\cellcolor{mathcolor}56.10 & \cellcolor{mathcolor}77.14 \\
    \textbf{Qwen2.5-Math} & Zero-shot & \cellcolor{mathcolor}82.50	&\cellcolor{mathcolor}85.50	&\cellcolor{mathcolor}92.30	&\cellcolor{mathcolor}53.60	&\cellcolor{mathcolor}55.40 & \cellcolor{mathcolor}73.86\\
    \hline
    \hline
    \multicolumn{8}{c}{\textit{\textbf{LLaMA3.1-8B}}} \\
    \hline
    \multirow{2}{*}{\textbf{Base LLM}} 
    & \cellcolor{graycolor}Zero-shot & \cellcolor{graycolor}63.60 & \cellcolor{graycolor}67.60 & \cellcolor{graycolor}73.40 & \cellcolor{graycolor}39.50 & \cellcolor{graycolor}31.70 & \cellcolor{graycolor}55.16 \\
    & Few-shot & 85.80 & 84.00 & 97.00 & 55.00 & 38.70 & 72.10 \\
    \cline{1-8}
    \multirow{2}{*}{\textbf{LLaMA}} 
    & \cellcolor{graycolor}CoT-SFT & \cellcolor{bettergreen}72.30 & \cellcolor{bettergreen}76.80 & \cellcolor{bettergreen}80.30 & \cellcolor{bettergreen}49.20 & \cellcolor{bettergreen}35.60 & \cellcolor{bettergreen}62.84 \\
    & FSLR & \cellcolor{bettergreen}\textbf{86.70} & \cellcolor{bettergreen}\textbf{82.40} & \cellcolor{bettergreen}\textbf{92.70} & \cellcolor{bettergreen}\textbf{52.00} & \cellcolor{bettergreen}\textbf{40.80} & \cellcolor{bettergreen}\textbf{70.92} \\
    \cline{1-8}
    \multirow{2}{*}{\textbf{Qwen}} 
    & \cellcolor{graycolor}CoT-SFT & \cellcolor{bettergreen}84.40 & \cellcolor{bettergreen}83.60 & \cellcolor{bettergreen}90.10 & \cellcolor{worsecolor}\textbf{70.00} & \cellcolor{bettergreen}43.40 & \cellcolor{bettergreen}74.30 \\
    & FSLR & \cellcolor{bettergreen}\textbf{87.80} & \cellcolor{bettergreen}\textbf{83.90} & \cellcolor{bettergreen}\textbf{93.80} & \cellcolor{worsecolor}67.00 & \cellcolor{bettergreen}\textbf{43.60} & \cellcolor{bettergreen}\textbf{75.22} \\
    \cline{1-8}
    \multirow{2}{*}{\textbf{Self}} 
    & \cellcolor{graycolor}CoT-SFT & \cellcolor{bettergreen}74.40 & \cellcolor{bettergreen}80.20 & \cellcolor{bettergreen}79.70 & \cellcolor{bettergreen}46.20 & \cellcolor{bettergreen}33.00 & \cellcolor{bettergreen}62.70 \\
    & FSLR & \cellcolor{bettergreen}\textbf{84.20} & \cellcolor{bettergreen}\textbf{80.90} & \cellcolor{bettergreen}\textbf{95.00} & \cellcolor{bettergreen}\textbf{53.70} & \cellcolor{bettergreen}\textbf{38.30} & \cellcolor{bettergreen}\textbf{70.42} \\
    \hline
    \hline
    \multicolumn{8}{c}{\textit{\textbf{Qwen2.5-7B}}} \\
    \hline
    \multirow{2}{*}{\textbf{Base LLM}} 
    & \cellcolor{graycolor}Zero-shot & \cellcolor{graycolor}84.20 & \cellcolor{graycolor}83.00 & \cellcolor{graycolor}90.80 & \cellcolor{graycolor}61.20 & \cellcolor{graycolor}53.40 & \cellcolor{graycolor}74.52 \\
    & Few-shot & 90.90 & 92.20 & 97.60 & 70.40 & 62.90 & 82.80 \\
    \cline{1-8}
    \multirow{2}{*}{\textbf{LLaMA}} 
    & \cellcolor{graycolor}CoT-SFT & \cellcolor{bettergreen}79.70 & \cellcolor{bettergreen}85.10 & \cellcolor{bettergreen}83.40 & \cellcolor{worsecolor}\textbf{49.00} & \cellcolor{bettergreen}50.70 & \cellcolor{bettergreen}69.58 \\
    & FSLR & \cellcolor{bettergreen}\textbf{88.60} & \cellcolor{bettergreen}\textbf{88.30} & \cellcolor{bettergreen}\textbf{94.10} & \cellcolor{worsecolor}47.20 & \cellcolor{bettergreen}\textbf{59.00} & \cellcolor{bettergreen}\textbf{75.46} \\
    \cline{1-8}
    \multirow{2}{*}{\textbf{Qwen}} 
    & \cellcolor{graycolor}CoT-SFT & \cellcolor{bettergreen}81.40 & \cellcolor{bettergreen}86.70 & \cellcolor{bettergreen}87.10 & \cellcolor{bettergreen}49.30 & \cellcolor{bettergreen}60.00 & \cellcolor{bettergreen}73.30 \\
    & FSLR & \cellcolor{bettergreen}\textbf{89.30} & \cellcolor{bettergreen}\textbf{87.40} & \cellcolor{bettergreen}\textbf{94.60} & \cellcolor{bettergreen}\textbf{52.10} & \cellcolor{bettergreen}\textbf{60.70} & \cellcolor{bettergreen}\textbf{76.82} \\
    \cline{1-8}
    \multirow{2}{*}{\textbf{Self}} 
    & \cellcolor{graycolor}CoT-SFT & \cellcolor{bettergreen}84.70 & \cellcolor{bettergreen}90.10 & \cellcolor{bettergreen}87.20 & \cellcolor{worsecolor}\textbf{47.70} & \cellcolor{bettergreen}61.50 & \cellcolor{bettergreen}74.24 \\
    & FSLR & \cellcolor{bettergreen}\textbf{85.60} & \cellcolor{bettergreen}\textbf{93.70} & \cellcolor{bettergreen}\textbf{90.10} & \cellcolor{worsecolor}45.80 & \cellcolor{bettergreen}\textbf{63.50} & \cellcolor{bettergreen}\textbf{75.74} \\
    \hline
    \hline
    \multicolumn{8}{c}{\textit{\textbf{Qwen3-4B}}} \\
    \hline
    \multirow{2}{*}{\textbf{Base LLM}} 
    & \cellcolor{graycolor}Zero-shot & \cellcolor{graycolor}78.10 & \cellcolor{graycolor}85.00 & \cellcolor{graycolor}88.10 & \cellcolor{graycolor}66.30 & \cellcolor{graycolor}62.40 & \cellcolor{graycolor}75.98 \\
    & Few-shot & 88.40 & 91.60 & 95.90 & 70.30 & 56.00 & 80.44 \\
    \cline{1-8}
    \multirow{2}{*}{\textbf{LLaMA}} 
    & \cellcolor{graycolor}CoT-SFT & \cellcolor{bettergreen}78.10 & \cellcolor{bettergreen}83.40 & \cellcolor{bettergreen}85.40 & \cellcolor{bettergreen}55.00 & \cellcolor{bettergreen}53.10 & \cellcolor{bettergreen}71.00 \\
    & FSLR & \cellcolor{bettergreen}\textbf{81.20} & \cellcolor{bettergreen}\textbf{84.20} & \cellcolor{bettergreen}\textbf{86.30} & \cellcolor{bettergreen}\textbf{67.20} & \cellcolor{bettergreen}\textbf{64.90} & \cellcolor{bettergreen}\textbf{76.76} \\
    \cline{1-8}
    \multirow{2}{*}{\textbf{Qwen}} 
    &\cellcolor{graycolor} CoT-SFT & \cellcolor{bettergreen}90.80 & \cellcolor{bettergreen}91.20 & \cellcolor{bettergreen}96.60 & \cellcolor{bettergreen}64.80 & \cellcolor{bettergreen}61.60 & \cellcolor{bettergreen}81.00 \\
    & FSLR & \cellcolor{bettergreen}\textbf{90.80} & \cellcolor{bettergreen}\textbf{93.80} & \cellcolor{bettergreen}\textbf{96.70} & \cellcolor{bettergreen}\textbf{69.10} & \cellcolor{bettergreen}\textbf{66.90} & \cellcolor{bettergreen}\textbf{83.34} \\
    \cline{1-8}
    \multirow{2}{*}{\textbf{Self}} 
    & \cellcolor{graycolor}CoT-SFT & \cellcolor{bettergreen}86.20 & \cellcolor{bettergreen}89.90 & \cellcolor{bettergreen}92.00 & \cellcolor{bettergreen}66.40 & \cellcolor{bettergreen}53.50 & \cellcolor{bettergreen}77.60 \\
    & FSLR & \cellcolor{bettergreen}\textbf{90.70} & \cellcolor{bettergreen}\textbf{93.50} & \cellcolor{bettergreen}\textbf{96.80} & \cellcolor{bettergreen}\textbf{69.60} & \cellcolor{bettergreen}\textbf{64.40} & \cellcolor{bettergreen}\textbf{83.00} \\
    \hline
  \end{tabular}
  }
  \caption{Out-of-distribution evaluation on five diverse benchmarks. Models are trained exclusively on GSM8K and evaluated on AsDIv, SVAMP, MAWPS, TabMWP, and GSM-Hard under zero-shot setting. Math-specialized models (DeepSeek-Math-7B and Qwen2.5-Math-7B) are evaluated zero-shot as reference baselines. \colorbox{bettergreen}{Green cells} indicate FSLR outperforms CoT-SFT. \colorbox{worsecolor}{Red cells} indicate FSLR underperforms CoT-SFT.}
  \label{tab:cross-domain-gsm8k}
\end{table*}

\subsection{Experiments and Analysis}
\label{sec:exp_results}

In this section, we evaluate FSLR from eight perspectives: in-distribution performance (Section~\ref{sec:in_dist}), out-of-distribution generalization (Section~\ref{sec:ood}), performance across problem complexity levels (Section~\ref{sec:complexity}), training efficiency (Section~\ref{sec:efficiency}), reliability of error attribution (Section~\ref{sec:reliability}), consistency analysis of teacher-generated first planning steps (Section~\ref{sec:consistency}), robustness to problem variations (Section~\ref{sec:robustness}), and case study (Section~\ref{sec:case_study}).
\subsubsection{In-Distribution Performance}
\label{sec:in_dist}

Table~\ref{tab:in-domain} presents results on in-distribution benchmarks 
(GSM8K and SVAMP), comparing FSLR against CoT-SFT using three teacher 
models for data generation.

\noindent\textbf{FSLR consistently outperforms CoT-SFT across all settings.} Across all three target models and teacher configurations, FSLR achieves superior or comparable performance to CoT-SFT, with particularly substantial improvements when using LLaMA-3.1-70B as the teacher: +5.2\% on LLaMA-3.1-8B (GSM8K), +5.5\% on LLaMA-3.1-8B (SVAMP), +7.5\% on Qwen2.5-7B (GSM8K), averaging +4.8\% improvement. When using Qwen2.5-72B and Self as data sources, FSLR achieves average improvements of +1.1\% and +3.8\% respectively. To further verify that FSLR's gains stem from the first-step design rather than general planning supervision, we compare against a Plan-and-Solve fine-tuning baseline~\cite{plan_tuning} in Appendix~\ref{plan_solve_baseline}.

\noindent\textbf{FSLR enables general-purpose models to surpass math-specialized 
models.} As shown in Table~\ref{tab:in-domain}, math-specialized models such as DeepSeek-Math-7B and Qwen2.5-Math-7B achieve 78.70\% and 79.20\% on GSM8K, and 82.20\% and 85.50\% on SVAMP respectively. In contrast, FSLR-trained general-purpose models substantially outperform these specialized baselines. For example, Qwen2.5-7B with self-generated FSLR data achieves 88.90\% on GSM8K and 93.00\% on SVAMP, surpassing Qwen2.5-Math-7B by +9.7\% and +7.5\% respectively. These gains suggest that models benefit more from focused supervision that isolates the understanding of logical relationships.

\noindent\textbf{FSLR demonstrates superior robustness to teacher model quality.} 
While CoT-SFT performance varies substantially across different teachers 
(82.28\% with LLaMA vs 87.80\% with Qwen, a 5.52\% gap), FSLR shows more 
consistent performance (87.08\% vs 88.90\%, only 1.82\% gap). Notably, even with self-generated data, FSLR achieves 88.48\% average accuracy, comparable to using larger teacher models. This robustness stems from the simplicity of first planning step supervision: generating a single logical planning decision is inherently easier and more reliable than generating complete solution trajectories, making FSLR less sensitive to teacher quality compared to CoT-SFT.

\paragraph{Comparison with prompting-based logical planning.} 
To evaluate whether prompting alone without fine-tuning can achieve similar benefits by instructing models to identify logical relationships before solving, we test two strategies: Zero-shot Logical Planning (Zero-LP) and Few-shot Logical Planning (Few-LP), which provides demonstrations of explicit logical relationship identification. As shown in Table~\ref{tab:in-domain}, both Zero-LP and Few-LP underperform FSLR by a considerable margin (82.45\% and 82.67\% vs. 87.08\%--88.90\%). This confirms that explicit logical relationship understanding requires internalization through fine-tuning rather than surface-level instruction following. Furthermore, Pass@$k$ evaluation (Appendix~\ref{pass_k}) confirms that FSLR expands model capability boundaries.

\subsubsection{Out-of-Distribution Performance}
\label{sec:ood}
Table~\ref{tab:cross-domain-gsm8k} presents out-of-distribution results 
where models are trained exclusively on GSM8K then evaluated on five 
diverse benchmarks: AsDIv, SVAMP, MAWPS, TabMWP, and GSM-Hard. This 
setup tests whether the improvements from FSLR training generalize 
beyond the training distribution.

\noindent\textbf{FSLR demonstrates superior generalization across diverse problem types.} FSLR consistently outperforms CoT-SFT across all three models and most evaluation datasets. On LLaMA-3.1-8B, FSLR achieves substantial improvements over CoT-SFT: +14.4\% on AsDIv (86.7\% vs 72.3\% with LLaMA teacher), +12.4\% on MAWPS, and +5.2\% on GSM-Hard, averaging +8.08\% improvement across all OOD datasets. Similarly, Qwen2.5-7B and Qwen3-4B achieve average gains of +3.52\% and +5.76\% respectively across teacher configurations. Consistent with in-distribution results, \textbf{FSLR-trained general-purpose models also surpass math-specialized models on OOD benchmarks}. This demonstrates that a more focused supervising logical relationship understanding leads to more transferable reasoning capabilities. We provide complementary Out--of-distribution experiments with models trained on SVAMP in Appendix~\ref{sec:ood-svamp}, which show consistent results.

% \noindent\textbf{FSLR enables general-purpose models to outperform 
% math-specialized models.} DeepSeek-Math-7B and Qwen2.5-Math-7B, despite 
% being specifically pre-trained for mathematical reasoning, achieve 77.14\% and 73.86\% average accuracy across the five OOD benchmarks respectively. In contrast, FSLR-trained general-purpose models surpass these specialized baselines: LLaMA-3.1-8B with Qwen teacher achieves 75.22\%, Qwen2.5-7B reaches 76.82\%, and Qwen3-4B achieves 83.34\%—outperforming DeepSeek-Math-7B by +6.2\%. 

\subsubsection{Performance Across Problem Complexity Levels}
\label{sec:complexity}
\begin{figure}[t]
  \includegraphics[width=\columnwidth]{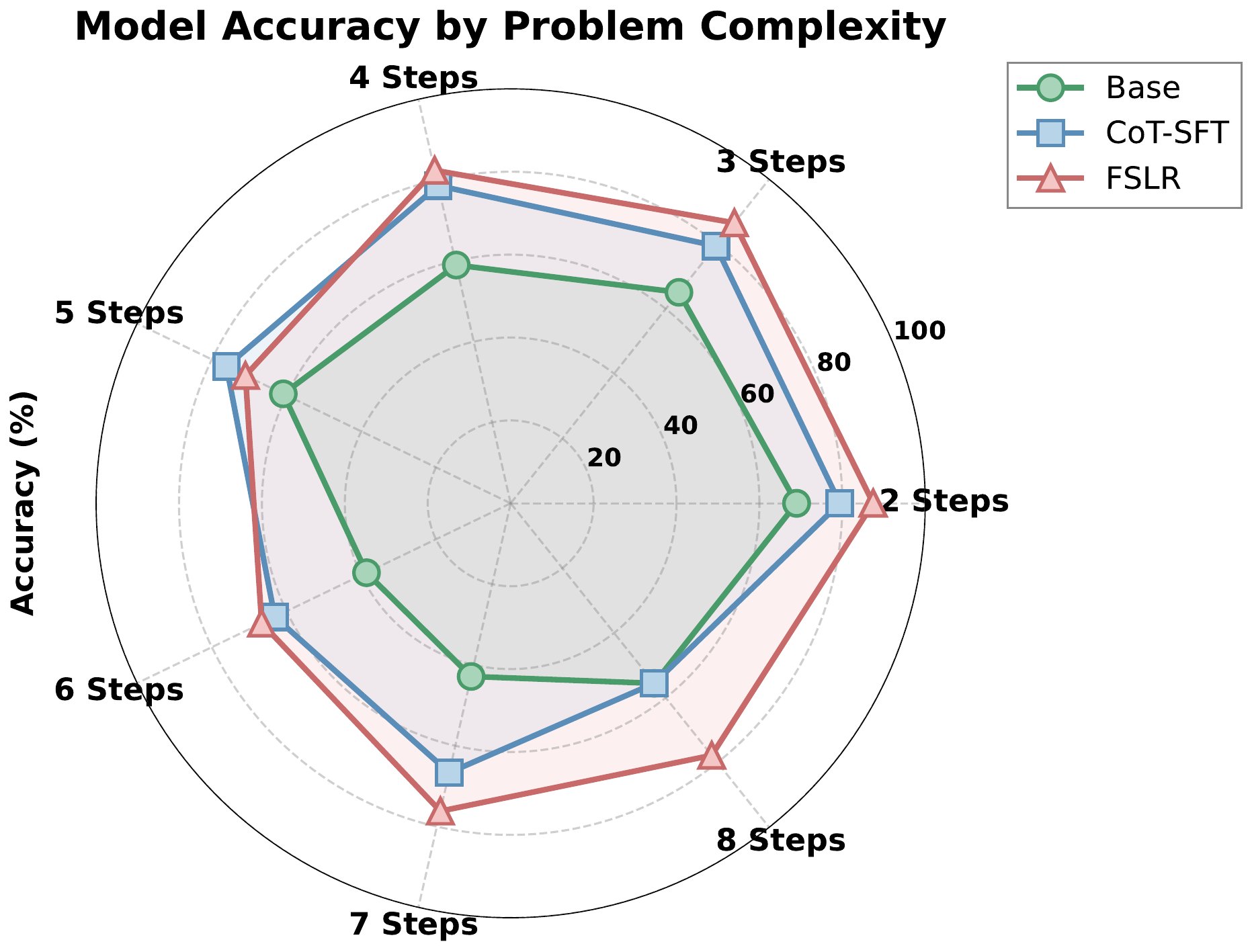}
  \caption{Radar chart showing model accuracy stratified by problem complexity (number of reasoning steps). Results shown are for LLaMA-3.1-8B with LLaMA-3.1-70B as the teacher model. FSLR consistently outperforms both Base and CoT-SFT methods, with particularly notable advantages on more complex problems requiring 6-8 reasoning steps.}
  \label{fig:different steps}
\end{figure}

To understand how FSLR's benefits vary with problem difficulty, we analyze model performance across problems requiring different numbers of reasoning steps (2-8 steps) on the GSM8K test set. Figure~\ref{fig:different steps} presents a radar chart comparing the base LLaMA-3.1-8B model, CoT-SFT, and FSLR across complexity levels. \textbf{FSLR demonstrates consistently superior performance across most complexity levels, with particularly strong advantages on challenging problems.} On simpler 2-4 step problems, FSLR achieves 82-87\% accuracy compared to 78-79\% for CoT-SFT and 59-69\% for the base model, representing +4-8\% improvements over CoT-SFT. Notably, on more complex 6-8 step problems where reasoning chains are much longer, FSLR shows dramatic gains compared to CoT-SFT: 67\% vs 63\% (6-step), 76\% vs 67\% (7-step), and 78\% vs 56\% (8-step), with the gap widening to +22\% on 8-step problems. This pattern reveals that FSLR's focused training on understanding logical relationships provides compounding benefits in multi-step reasoning.

\subsubsection{Training Efficiency}
\label{sec:efficiency}

\begin{table*}[t]
  \centering
  \normalsize
  % \setlength{\tabcolsep}{6pt} 
  % Add these to preamble:
  % \usepackage{colortbl}
  % \usepackage{xcolor}
  % \definecolor{bettergreen}{rgb}{0.85, 0.95, 0.85}
  % \definecolor{worsecolor}{rgb}{1.0, 0.85, 0.85}
  \begin{tabular}{l|l|cc|cc|cc}
    \hline
    \multirow{2}{*}{\textbf{Token length}} & \multirow{2}{*}{\textbf{Data Source}} & \multicolumn{2}{c|}{\textit{\textbf{LLaMA3.1-8B}}} & \multicolumn{2}{c|}{\textit{\textbf{Qwen2.5-7B}}} & \multicolumn{2}{c}{\textit{\textbf{Qwen3-4B}}} \\
    \cline{3-8}
    & & \cellcolor{graycolor}\textbf{CoT-SFT} & \cellcolor{graycolor}\textbf{FSLR} & \cellcolor{graycolor}\textbf{CoT-SFT} & \cellcolor{graycolor}\textbf{FSLR} & \cellcolor{graycolor}\textbf{CoT-SFT} & \cellcolor{graycolor}\textbf{FSLR} \\
    \hline
    \hline
    \multirow{3}{*}{\textbf{GSM8K}} 
    & \textbf{LLaMA}  & \cellcolor{bettergreen}237.64 & \cellcolor{bettergreen}\textbf{37.82} & \cellcolor{bettergreen}254.98 & \cellcolor{bettergreen}\textbf{38.12} & \cellcolor{bettergreen}254.98 & \cellcolor{bettergreen}\textbf{38.12} \\
    & \cellcolor{graycolor}\textbf{Qwen}  & \cellcolor{bettergreen}293.76 & \cellcolor{bettergreen}\textbf{38.95} & \cellcolor{bettergreen}310.38 & \cellcolor{bettergreen}\textbf{39.24} & \cellcolor{bettergreen}310.38 & \cellcolor{bettergreen}\textbf{39.24} \\
    & \textbf{Self}  & \cellcolor{bettergreen}256.41 & \cellcolor{bettergreen}\textbf{27.02} & \cellcolor{bettergreen}301.22 & \cellcolor{bettergreen}\textbf{51.92} & \cellcolor{bettergreen}262.11 & \cellcolor{bettergreen}\textbf{42.37} \\
    \hline
    \multirow{3}{*}{\textbf{SVAMP}} 
    & \textbf{LLaMA} & \cellcolor{bettergreen}195.51 & \cellcolor{bettergreen}\textbf{40.90} & \cellcolor{bettergreen}207.43 & \cellcolor{bettergreen}\textbf{41.42} & \cellcolor{bettergreen}207.43 & \cellcolor{bettergreen}\textbf{41.42} \\
    & \cellcolor{graycolor}\textbf{Qwen}  & \cellcolor{bettergreen}206.19 & \cellcolor{bettergreen}\textbf{37.11} & \cellcolor{bettergreen}215.12 & \cellcolor{bettergreen}\textbf{37.66} & \cellcolor{bettergreen}215.12 & \cellcolor{bettergreen}\textbf{37.66} \\
    & \textbf{Self}  & \cellcolor{bettergreen}244.67 & \cellcolor{bettergreen}\textbf{36.56} & \cellcolor{bettergreen}217.36 & \cellcolor{bettergreen}\textbf{33.55} & \cellcolor{bettergreen}192.36 & \cellcolor{bettergreen}\textbf{45.98} \\
    \hline
  \end{tabular}
  \caption{Average token length of training sequences for CoT-SFT and FSLR across different models and datasets. \colorbox{bettergreen}{Green cells} indicate FSLR uses fewer tokens than CoT-SFT (lower is better for efficiency).}
  \label{tab:token-length}
\end{table*}

Beyond improved accuracy, \textbf{FSLR offers substantial computational advantages over CoT-SFT.} Table~\ref{tab:token-length} shows that FSLR training data contains significantly fewer tokens: on GSM8K, FSLR produces sequences averaging 27–52 tokens, compared to 238–310 tokens for CoT-SFT (an 84-87\% reduction), while on SVAMP, the reduction is similarly dramatic (34-46 tokens vs. 192-245 tokens, 81-86\% reduction). This compression stems from FSLR's focus on first planning step rather than complete solution trajectories, eliminating the extensive computational execution steps that dominate CoT sequences.

\begin{figure*}[h]
  \centering
  \includegraphics[width=0.95\textwidth]{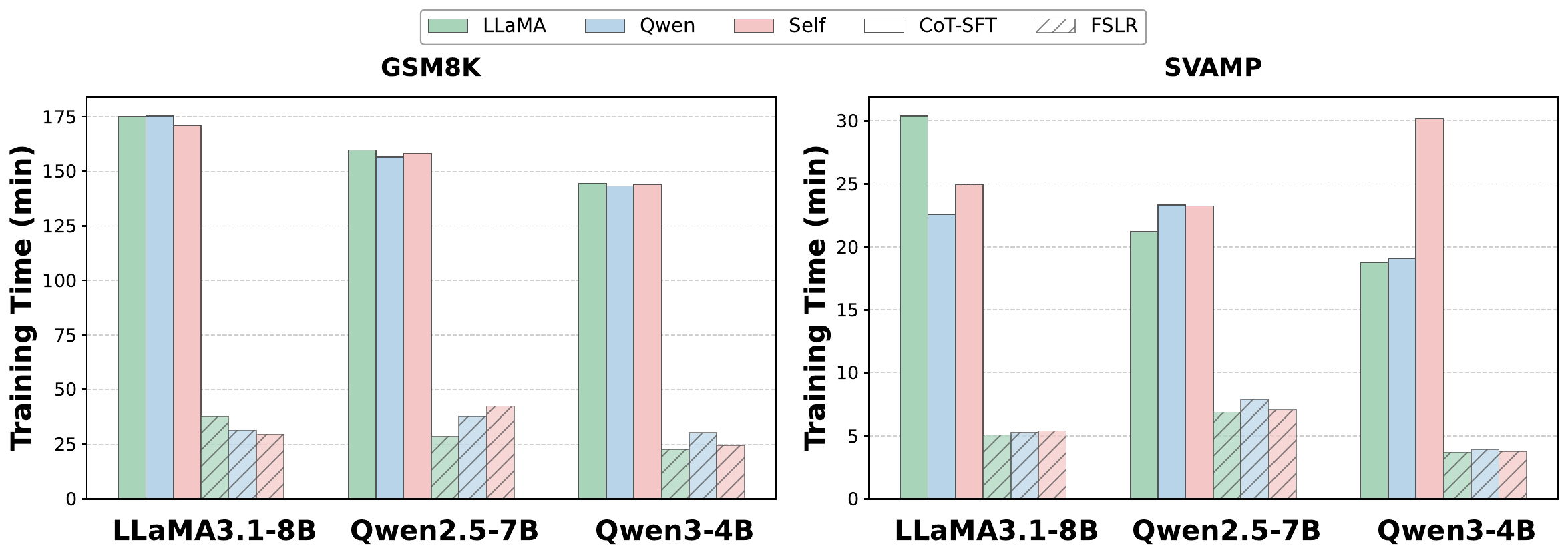}
  \caption{Training time (in minutes) for CoT-SFT and FSLR across different models and data sources. FSLR achieves substantial speedup over CoT-SFT on both GSM8K (left) and SVAMP (right), reducing training time by approximately 4-6$\times$ while maintaining competitive performance.}
  \label{fig:training time}
\end{figure*}

\noindent Figure~\ref{fig:training time} demonstrates that these token savings directly translate to training acceleration. In GSM8K, the FSLR training is completed in 23-43 minutes compared to 144-175 minutes for CoT-SFT, achieving 4-6× speedup across all models and teacher configurations. Similar gains are observed on SVAMP (4–6× speedup), demonstrating that FSLR's training is not only more effective but also substantially more efficient.

% than training on complete reasoning trajectories. 

\begin{figure}[t]
  \includegraphics[width=0.95\columnwidth]{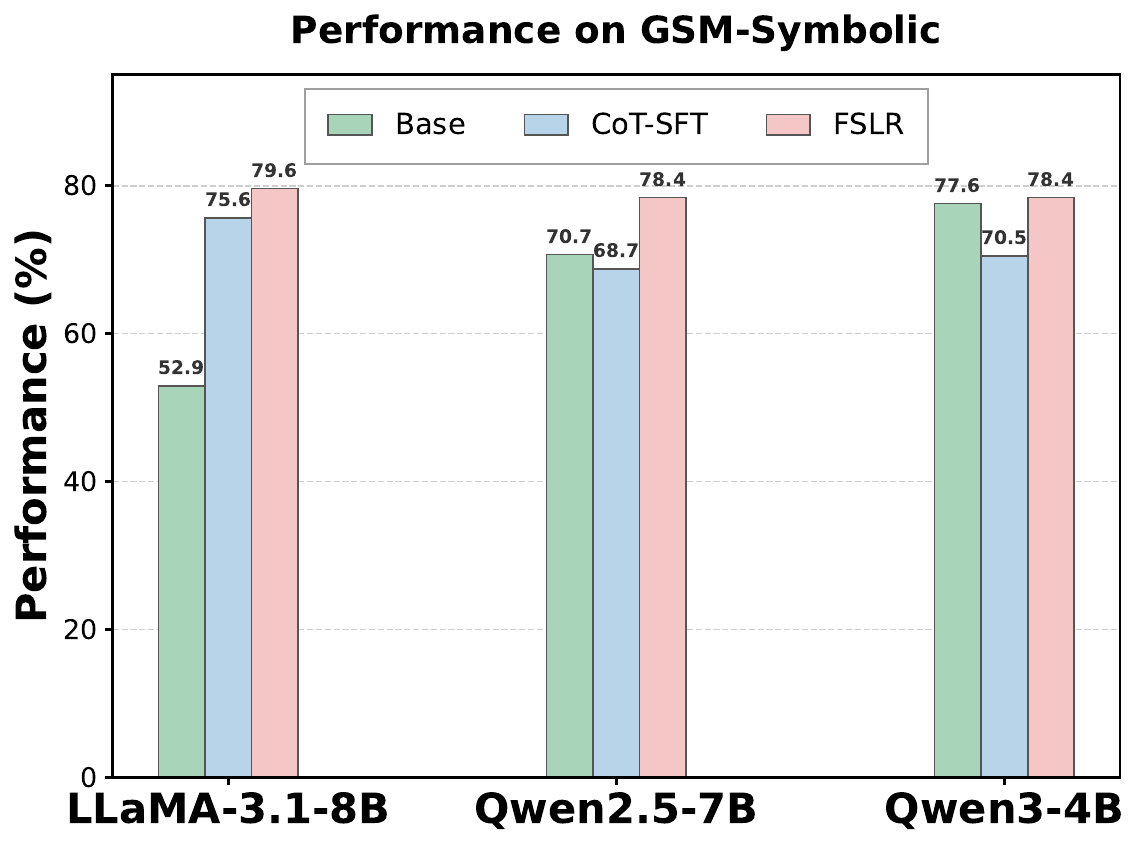}
  \caption{Zero-shot performance on GSM-Symbolic benchmark. All models are trained using LLaMA-3.1-70B as the teacher model. FSLR achieves the best performance across all three models, demonstrating superior generalization.}
  \label{fig:gsm_symbolic}
\end{figure}

\subsubsection{Reliability of Error Attribution}
\label{sec:reliability}

To validate the robustness of error attribution, we conduct additional analysis using LLaMA-3.1-70B-Instruct and Qwen2.5-72B-Instruct as independent judges with the same classification prompt (Appendix~\ref{multi_judge}). As shown in Table~\ref{tab:judge}, logical relationship understanding errors consistently account for over 90\% of failures across all three judges, models, and methods. The convergence across judges with different architectures and training backgrounds substantially strengthens the validity of our error attribution. Furthermore, manual inspection of 100 randomly sampled cases confirms that judge classifications align with human judgment at over 95\% accuracy.

\subsubsection{Consistency Analysis of Teacher-Generated First Planning Steps}
\label{sec:consistency}

To investigate whether potential multiple first steps for a given problem 
introduce noise or limit the student model, we conduct a two-part consistency analysis on 500 randomly sampled problems from the GSM8K training set, using GPT-4o to judge semantic consistency of generated first planning steps.

\noindent\textbf{Within-teacher consistency.} We generate first planning steps using LLaMA-3.1-70B-Instruct at three temperatures (0, 0.5, 1.0) and evaluate whether outputs are semantically consistent across temperature settings. Results show that 84.7\% of problems yield consistent first planning steps across all three temperatures, indicating that the first planning step is generally well-defined and robust to generation variance.

\noindent\textbf{Across-teacher consistency.} We further analyze whether different teacher models produce consistent first planning steps by comparing outputs from LLaMA-3.1-70B-Instruct, Qwen2.5-72B-Instruct, and the self-generated condition. We find that 75.4\% of problems show consistent first planning steps across all three teachers. The 9.3\% gap between within- and across-teacher consistency is expected given differences in model capabilities and training backgrounds, yet the majority of problems still yield consistent logical understanding across diverse teacher sources.

\noindent Together, the high consistency rates both within-teacher (84.7\%) and across-teacher (75.4\%) demonstrate that the first planning step is generally well-defined for mathematical problems. 

\subsubsection{Performance on GSM-Symbolic}
\label{sec:robustness}

\noindent To further validate FSLR's robustness beyond standard benchmarks, we evaluate on GSM-Symbolic~\cite{gsm_symbolic}, a benchmark specifically designed to test genuine reasoning.

\noindent Figure~\ref{fig:gsm_symbolic} presents performance across all three models on GSM-Symbolic. \textbf{FSLR demonstrates superior robustness to problem variations.} While all models show performance drops compared to standard GSM8K, FSLR maintains substantially stronger performance than both base models and CoT-SFT. On LLaMA-3.1-8B, FSLR achieves 79.6\% accuracy compared to 70.7\% for CoT-SFT and 52.9\% for the base model, achieving +8.9\% improvement over CoT-SFT. Similarly, FSLR achieves improvements of +9.7\% on Qwen2.5-7B (78.4\% vs 68.7\%) and +7.9\% on Qwen3-4B (78.4\% vs 70.5\%) over CoT-SFT. This consistent advantage across different models suggests that FSLR's focused training on understanding logical relationships develops more robust reasoning capabilities, indicating that models trained with FSLR rely less on memorized patterns and more on genuine understanding of operational dependencies, which is the capability GSM-Symbolic is designed to test.

\subsubsection{Case Study}
\label{sec:case_study}

To illustrate how FSLR improves logical relationship understanding, we present an example from GSM8K where the base model and CoT-SFT both fail, while FSLR succeeds. The following example uses LLaMA-3.1-8B as the target model with LLaMA-3.1-70B as the teacher model. Complete reasoning content is provided in Appendix~\ref{sec:case-study}.

\noindent\textbf{Problem:} \textit{Shiela bought 5 cell phones for \$150 each for a 3-month installment. A 2\% interest will be charged for each unit. How much will Shiela pay each month for 3 months?}

\noindent\textbf{Ground Truth:} \textit{A \$150 × 2/100 = \$3 interest will be charged for each unit. So, each cell phone costs \$150 + \$3 = \$153. The total amount to be paid for 5 cell phones is \$153 × 5 = \$765. Thus, Shiela will pay \$765/3 = \$255 each month.} \textbf{Answer: \$255}

\noindent\textbf{Base LLM:} \textit{...Interest = 2\% of \$150 = \$3. \underline{Total interest = 3 × \$3 = \$9}. Total amount = \$750 + \$9 = \$759. Monthly payment = \$759/3 = \$253.} \textbf{Answer: \$253} 

\noindent\textbf{CoT-SFT:} \textit{...The interest rate is 2\% per unit, and there are 5 units. \underline{So, the total interest charged is: 5 × 2\% = 10\%}. Total = \$750 + \$750 × 10\% = \$825. Monthly = \$825/3 = \$275.} \textbf{Answer: \$275} 

\noindent\textbf{FSLR:} \textit{...Total cost = 5 × \$150 = \$750. \underline{Total interest: 2\% of \$750 = 0.02 × \$750 = \$15}. Total amount = \$750 + \$15 = \$765. Monthly payment = \$765/3 = \$255.} \textbf{Answer: \$255} 

\noindent Both the base LLM and CoT-SFT fail because they misidentify which quantity the interest rate should be applied to. The base model incorrectly interprets "interest for each unit" as applying to the payment period, computing "3 months × \$3 = \$9" total interest. CoT-SFT makes a different error: it aggregates the interest rate itself ("5 × 2\% = 10\%"), then applies this to the total cost, completely missing that "per unit" means the 2\% must be calculated on the \textit{individual unit price} of \$150. In contrast, FSLR correctly identifies the operational structure, recognizing that the problem requires calculating "2\% of \$750", which properly captures the relationship between the interest rate, unit price, and quantity. Notably, all three models demonstrate sound arithmetic execution: the computational steps are performed correctly given their operational decisions.

\section{Conclusion}

In this work, we identified that logical relationship understanding errors constitute the primary bottleneck in mathematical reasoning, accounting for over 90\% of incorrect predictions, and Chain-of-Thought Supervised Fine-Tuning(CoT-SFT) fails to substantially address this limitation. To bridge this gap, we proposed \textbf{F}irst-\textbf{S}tep \textbf{L}ogical \textbf{R}easoning (\textbf{FSLR}), a lightweight framework that provides a more focused training signal for logical relationship understanding by training models to identify the first planning step. Extensive experiments demonstrate that FSLR consistently outperforms CoT-SFT across multiple models and benchmarks, achieving stronger generalization on out-of-distribution tasks while requiring 81-87\% fewer training tokens. 
%  Our findings suggest that a more focused supervising logical relationship understanding yields more transferable reasoning capabilities than training on complete solution trajectories, opening promising directions for improving reasoning capabilities in LLMs.
\section{Limitations}
Our work has several limitations. First, FSLR is evaluated on mathematical problems, and its effectiveness on other reasoning domains remains unexplored. Second, our framework relies on supervised fine-tuning with teacher-generated annotations, which may limit the model's ability. Exploring reinforcement learning approaches that reward correct logical relationship identification could potentially yield further improvements in logical relationship understanding and is left for future work.
% \bibliography{anthology,cu}
\bibliography{main}

@article{reasoning_survey,
  title={A survey of reasoning with foundation models: Concepts, methodologies, and outlook},
  author={Sun, Jiankai and Zheng, Chuanyang and Xie, Enze and Liu, Zhengying and Chu, Ruihang and Qiu, Jianing and Xu, Jiaqi and Ding, Mingyu and Li, Hongyang and Geng, Mengzhe and others},
  journal={ACM Computing Surveys},
  volume={57},
  number={11},
  pages={1--43},
  year={2025},
  publisher={ACM New York, NY}
}

@article{sft_memorize,
  title={Sft memorizes, rl generalizes: A comparative study of foundation model post-training},
  author={Chu, Tianzhe and Zhai, Yuexiang and Yang, Jihan and Tong, Shengbang and Xie, Saining and Schuurmans, Dale and Le, Quoc V and Levine, Sergey and Ma, Yi},
  journal={arXiv preprint arXiv:2501.17161},
  year={2025}
}

@article{gpt4o,
  title={Gpt-4o system card},
  author={Hurst, Aaron and Lerer, Adam and Goucher, Adam P and Perelman, Adam and Ramesh, Aditya and Clark, Aidan and Ostrow, AJ and Welihinda, Akila and Hayes, Alan and Radford, Alec and others},
  journal={arXiv preprint arXiv:2410.21276},
  year={2024}
}

@article{gsm_symbolic,
  title={Gsm-symbolic: Understanding the limitations of mathematical reasoning in large language models},
  author={Mirzadeh, Iman and Alizadeh, Keivan and Shahrokhi, Hooman and Tuzel, Oncel and Bengio, Samy and Farajtabar, Mehrdad},
  journal={arXiv preprint arXiv:2410.05229},
  year={2024}
}

@article{self_explore_sft,
  title={Self-explore: Enhancing mathematical reasoning in language models with fine-grained rewards},
  author={Hwang, Hyeonbin and Kim, Doyoung and Kim, Seungone and Ye, Seonghyeon and Seo, Minjoon},
  journal={arXiv preprint arXiv:2404.10346},
  year={2024}
}

@article{teaching_to_small_llm,
  title={Teaching arithmetic to small transformers},
  author={Lee, Nayoung and Sreenivasan, Kartik and Lee, Jason D and Lee, Kangwook and Papailiopoulos, Dimitris},
  journal={arXiv preprint arXiv:2307.03381},
  year={2023}
}

@article{learning_composable_cot,
  title={Learning Composable Chains-of-Thought},
  author={Yin, Fangcong and Liu, Zeyu Leo and Leqi, Liu and Ye, Xi and Durrett, Greg},
  journal={arXiv preprint arXiv:2505.22635},
  year={2025}
}

@inproceedings{self_train_meets_consistency,
  title={Self-training meets consistency: Improving llms’ reasoning with consistency-driven rationale evaluation},
  author={Lee, Jaehyeok and Sakaguchi, Keisuke and Bak, JinYeong},
  booktitle={Proceedings of the 2025 Conference of the Nations of the Americas Chapter of the Association for Computational Linguistics: Human Language Technologies (Volume 1: Long Papers)},
  pages={10519--10539},
  year={2025}
}

@inproceedings{impact_cot_sft,
  title={On the impact of fine-tuning on chain-of-thought reasoning},
  author={Lobo, Elita and Agarwal, Chirag and Lakkaraju, Himabindu},
  booktitle={Proceedings of the 2025 Conference of the Nations of the Americas Chapter of the Association for Computational Linguistics: Human Language Technologies (Volume 1: Long Papers)},
  pages={11679--11698},
  year={2025}
}

@inproceedings{llm_mimic_human,
  title={How likely do llms with cot mimic human reasoning?},
  author={Bao, Guangsheng and Zhang, Hongbo and Wang, Cunxiang and Yang, Linyi and Zhang, Yue},
  booktitle={Proceedings of the 31st International Conference on Computational Linguistics},
  pages={7831--7850},
  year={2025}
}

@article{gsm8k,
  title={Training verifiers to solve math word problems},
  author={Cobbe, Karl and Kosaraju, Vineet and Bavarian, Mohammad and Chen, Mark and Jun, Heewoo and Kaiser, Lukasz and Plappert, Matthias and Tworek, Jerry and Hilton, Jacob and Nakano, Reiichiro and others},
  journal={arXiv preprint arXiv:2110.14168},
  year={2021}
}

@article{svamp,
  title={Are NLP models really able to solve simple math word problems?},
  author={Patel, Arkil and Bhattamishra, Satwik and Goyal, Navin},
  journal={arXiv preprint arXiv:2103.07191},
  year={2021}
}

@inproceedings{asdiv,
  title={A diverse corpus for evaluating and developing English math word problem solvers},
  author={Miao, Shen-Yun and Liang, Chao-Chun and Su, Keh-Yih},
  booktitle={Proceedings of the 58th annual meeting of the Association for Computational Linguistics},
  pages={975--984},
  year={2020}
}

@inproceedings{mawps,
  title={MAWPS: A math word problem repository},
  author={Koncel-Kedziorski, Rik and Roy, Subhro and Amini, Aida and Kushman, Nate and Hajishirzi, Hannaneh},
  booktitle={Proceedings of the 2016 conference of the north american chapter of the association for computational linguistics: human language technologies},
  pages={1152--1157},
  year={2016}
}

@article{tabmwp,
  title={Dynamic prompt learning via policy gradient for semi-structured mathematical reasoning},
  author={Lu, Pan and Qiu, Liang and Chang, Kai-Wei and Wu, Ying Nian and Zhu, Song-Chun and Rajpurohit, Tanmay and Clark, Peter and Kalyan, Ashwin},
  journal={arXiv preprint arXiv:2209.14610},
  year={2022}
}

@article{gsm_hard,
  title={PAL: Program-aided Language Models},
  author={Gao, Luyu and Madaan, Aman and Zhou, Shuyan and Alon, Uri and Liu, Pengfei and Yang, Yiming and Callan, Jamie and Neubig, Graham},
  journal={arXiv preprint arXiv:2211.10435},
  year={2022}
}

@article{llama3.1,
  title={The llama 3 herd of models},
  author={Grattafiori, Aaron and Dubey, Abhimanyu and Jauhri, Abhinav and Pandey, Abhinav and Kadian, Abhishek and Al-Dahle, Ahmad and Letman, Aiesha and Mathur, Akhil and Schelten, Alan and Vaughan, Alex and others},
  journal={arXiv preprint arXiv:2407.21783},
  year={2024}
}

@article{qwen3,
  title={Qwen3 technical report},
  author={Yang, An and Li, Anfeng and Yang, Baosong and Zhang, Beichen and Hui, Binyuan and Zheng, Bo and Yu, Bowen and Gao, Chang and Huang, Chengen and Lv, Chenxu and others},
  journal={arXiv preprint arXiv:2505.09388},
  year={2025}
}

@article{qwen2,
  title={Qwen2 technical report},
  author={Team, Qwen and others},
  journal={arXiv preprint arXiv:2407.10671},
  volume={2},
  number={3},
  year={2024}
}

@article{llm4mathreasoning,
  title={Large language models for mathematical reasoning: Progresses and challenges},
  author={Ahn, Janice and Verma, Rishu and Lou, Renze and Liu, Di and Zhang, Rui and Yin, Wenpeng},
  journal={arXiv preprint arXiv:2402.00157},
  year={2024}
}

@article{llm_mathematical_optimization_survey,
  title={A survey on mathematical reasoning and optimization with large language models},
  author={Forootani, Ali},
  journal={arXiv preprint arXiv:2503.17726},
  year={2025}
}

@article{llm_mathematical_survey,
  title={A Survey on Large Language Models for Mathematical Reasoning},
  author={Wang, Peng-Yuan and Liu, Tian-Shuo and Wang, Chenyang and Wang, Yi-Di and Yan, Shu and Jia, Cheng-Xing and Liu, Xu-Hui and Chen, Xin-Wei and Xu, Jia-Cheng and Li, Ziniu and others},
  journal={arXiv preprint arXiv:2506.08446},
  year={2025}
}

@article{gsm_plus,
  title={Gsm-plus: A comprehensive benchmark for evaluating the robustness of llms as mathematical problem solvers},
  author={Li, Qintong and Cui, Leyang and Zhao, Xueliang and Kong, Lingpeng and Bi, Wei},
  journal={arXiv preprint arXiv:2402.19255},
  year={2024}
}

@article{math_perturb,
  title={Math-perturb: Benchmarking llms’ math reasoning abilities against hard perturbations, 2025},
  author={Huang, Kaixuan and Guo, Jiacheng and Li, Zihao and Ji, Xiang and Ge, Jiawei and Li, Wenzhe and Guo, Yingqing and Cai, Tianle and Yuan, Hui and Wang, Runzhe and others},
  journal={URL https://arxiv. org/abs/2502.06453}
}

@inproceedings{gsm_ic,
  title={Large language models can be easily distracted by irrelevant context},
  author={Shi, Freda and Chen, Xinyun and Misra, Kanishka and Scales, Nathan and Dohan, David and Chi, Ed H and Sch{\"a}rli, Nathanael and Zhou, Denny},
  booktitle={International Conference on Machine Learning},
  pages={31210--31227},
  year={2023},
  organization={PMLR}
}

@article{logical_reasoning,
author = {Serna M., Edgar and Serna, Alexei},
year = {2015},
month = {08},
pages = {325-331},
title = {Knowledge in Engineering: A View from the Logical Reasoning},
volume = {7},
journal = {International Journal of Computer Theory and Engineering},
doi = {10.7763/IJCTE.2015.V7.980}
}

@article{qwen2.5_math,
  title={Qwen2.5-Math Technical Report: Toward Mathematical Expert Model via Self-Improvement}, 
  author={An Yang and Beichen Zhang and Binyuan Hui and Bofei Gao and Bowen Yu and Chengpeng Li and Dayiheng Liu and Jianhong Tu and Jingren Zhou and Junyang Lin and Keming Lu and Mingfeng Xue and Runji Lin and Tianyu Liu and Xingzhang Ren and Zhenru Zhang},
  journal={arXiv preprint arXiv:2409.12122},
  year={2024}
}

@article{deepseek_math,
  title={Deepseekmath: Pushing the limits of mathematical reasoning in open language models},
  author={Shao, Zhihong and Wang, Peiyi and Zhu, Qihao and Xu, Runxin and Song, Junxiao and Bi, Xiao and Zhang, Haowei and Zhang, Mingchuan and Li, YK and Wu, Yang and others},
  journal={arXiv preprint arXiv:2402.03300},
  year={2024}
}

@inproceedings{plan_tuning,
  title={Plan-tuning: Post-training language models to learn step-by-step planning for complex problem solving},
  author={Parmar, Mihir and Goyal, Palash and Liu, Xin and Song, Yiwen and Ling, Mingyang and Baral, Chitta and Palangi, Hamid and Pfister, Tomas},
  booktitle={Proceedings of the 2025 Conference on Empirical Methods in Natural Language Processing},
  pages={21430--21444},
  year={2025}
}

\appendix

\begin{table*}[h]
  \centering
  \normalsize
  \setlength{\tabcolsep}{4pt}
  \resizebox{0.95\textwidth}{!}{%
  \begin{tabular}{l|l|cc|cc|cc|c}
    \hline
    \multicolumn{2}{c|}{\cellcolor{graycolor}\textbf{In-Distribution Results}} & 
    \multicolumn{2}{c|}{\cellcolor{graycolor}\textit{\textbf{LLaMA3.1-8B}}} & 
    \multicolumn{2}{c|}{\cellcolor{graycolor}\textit{\textbf{Qwen2.5-7B}}} & 
    \multicolumn{2}{c|}{\cellcolor{graycolor}\textit{\textbf{Qwen3-4B}}} & 
    \multirow{2}{*}{\textbf{Average}} \\
    \cline{1-8}
    \textbf{Data Source} & \textbf{Method} & 
    \textbf{GSM8K} & \textbf{SVAMP} & 
    \textbf{GSM8K} & \textbf{SVAMP} & 
    \textbf{GSM8K} & \textbf{SVAMP} & \\
    \hline
    \hline
    \multirow{3}{*}{\textbf{LLaMA}} 
    & CoT-SFT 
        & \cellcolor{bettergreen}77.90 
        & \cellcolor{bettergreen}79.30 
        & \cellcolor{bettergreen}77.60 
        & \cellcolor{bettergreen}88.10 
        & \cellcolor{bettergreen}85.70 
        & \cellcolor{bettergreen}85.10 
        & \cellcolor{bettergreen}82.28 \\
    & Plan-and-Solve 
        & \cellcolor{bettergreen}81.70 
        & \cellcolor{bettergreen}83.30 
        & \cellcolor{bettergreen}83.70 
        & \cellcolor{bettergreen}89.00 
        & \cellcolor{bettergreen}\textbf{87.50} 
        & \cellcolor{bettergreen}90.10 
        & \cellcolor{bettergreen}85.88 \\
    & \cellcolor{graycolor}FSLR 
        & \cellcolor{bettergreen}\textbf{83.10} 
        & \cellcolor{bettergreen}\textbf{84.80} 
        & \cellcolor{bettergreen}\textbf{85.10} 
        & \cellcolor{bettergreen}\textbf{91.30} 
        & \cellcolor{bettergreen}87.10 
        & \cellcolor{bettergreen}\textbf{91.10} 
        & \cellcolor{bettergreen}\textbf{87.08} \\
    \hline
  \end{tabular}
  }
  \caption{Comparison with Plan-and-Solve fine-tuning on in-distribution benchmarks GSM8K and SVAMP. 
  All models use LLaMA-3.1-70B-Instruct as the teacher model. 
  \colorbox{bettergreen}{Green cells} indicate the method outperforms CoT-SFT.}
  \label{tab:plan-and-solve}
\end{table*}

\begin{table*}[!h]
  \centering
  \normalsize
  \setlength{\tabcolsep}{6pt}% 减少列间距以适应页面宽度
  % Add these to preamble:
  % \usepackage{colortbl}
  % \usepackage{xcolor}
  % \definecolor{bettergreen}{rgb}{0.85, 0.95, 0.85}
  % \definecolor{worsecolor}{rgb}{1.0, 0.85, 0.85}
  \begin{tabular}{l|l|ccccc|c}
    \hline
    \multicolumn{2}{c|}{\cellcolor{graycolor}\textbf{Out-of-Distribution Results}} & \multicolumn{5}{c|}{\cellcolor{graycolor}\textbf{Models Trained on SVAMP}} & \multirow{2}{*}{\textbf{Average}} \\
    \cline{1-7}
    \textbf{Data Source} & \textbf{Method} & \textbf{AsDIv} & \textbf{GSM8K} & \textbf{MAWPS} & \textbf{TabMWP} & \textbf{GSM-Hard} & \\
    \hline
    \hline
    \multicolumn{8}{c}{\textit{\textbf{Math-Specialized Models}}} \\
    \hline
    \cellcolor{graycolor}\textbf{DeepSeek-Math} & \cellcolor{graycolor}Zero-shot & \cellcolor{mathcolor}85.00 & \cellcolor{mathcolor}82.20 & \cellcolor{mathcolor}92.50 & \cellcolor{mathcolor}69.90 & \cellcolor{mathcolor}56.10 & \cellcolor{mathcolor}77.14 \\
    \textbf{Qwen2.5-Math} & Zero-shot & \cellcolor{mathcolor}82.50 & \cellcolor{mathcolor}85.50 & \cellcolor{mathcolor}92.30 & \cellcolor{mathcolor}53.60 & \cellcolor{mathcolor}55.40 & \cellcolor{mathcolor}73.86 \\
    \hline
    \hline
    \multicolumn{8}{c}{\textit{\textbf{LLaMA3.1-8B}}} \\
    \hline
    \multirow{2}{*}{\textbf{Base LLM}} 
    & \cellcolor{graycolor}Zero-shot & \cellcolor{graycolor}63.60 & \cellcolor{graycolor}62.90 & \cellcolor{graycolor}73.40 & \cellcolor{graycolor}39.50 & \cellcolor{graycolor}31.70 & \cellcolor{graycolor}54.22 \\
    & Few-shot & 85.80 & 77.50 & 97.00 & 55.00 & 38.70 & 70.80 \\
    \cline{1-8}
    \multirow{2}{*}{\textbf{LLaMA}} 
    & \cellcolor{graycolor}CoT-SFT & \cellcolor{bettergreen}75.70 & \cellcolor{bettergreen}77.40 & \cellcolor{bettergreen}82.10 & \cellcolor{bettergreen}47.10 & \cellcolor{bettergreen}34.00 & \cellcolor{bettergreen}63.26 \\
    & FSLR & \cellcolor{bettergreen}\textbf{85.90} & \cellcolor{bettergreen}\textbf{83.20} & \cellcolor{bettergreen}\textbf{94.40} & \cellcolor{bettergreen}\textbf{55.70} & \cellcolor{bettergreen}\textbf{41.70} & \cellcolor{bettergreen}\textbf{72.18} \\
    \cline{1-8}
    \multirow{2}{*}{\textbf{Qwen}} 
    & \cellcolor{graycolor}CoT-SFT & \cellcolor{bettergreen}77.40 & \cellcolor{bettergreen}74.70 & \cellcolor{bettergreen}84.30 & \cellcolor{bettergreen}54.30 & \cellcolor{bettergreen}36.80 & \cellcolor{bettergreen}65.50 \\
    & FSLR & \cellcolor{bettergreen}\textbf{79.90} & \cellcolor{bettergreen}\textbf{79.60} & \cellcolor{bettergreen}\textbf{94.20} & \cellcolor{bettergreen}\textbf{59.00} & \cellcolor{bettergreen}\textbf{39.20} & \cellcolor{bettergreen}\textbf{70.38} \\
    \cline{1-8}
    \multirow{2}{*}{\textbf{Self}} 
    & \cellcolor{graycolor}CoT-SFT & \cellcolor{bettergreen}73.60 & \cellcolor{bettergreen}75.40 & \cellcolor{bettergreen}78.10 & \cellcolor{bettergreen}40.40 & \cellcolor{bettergreen}35.90 & \cellcolor{bettergreen}60.68 \\
    & FSLR & \cellcolor{bettergreen}\textbf{83.30} & \cellcolor{bettergreen}\textbf{82.00} & \cellcolor{bettergreen}\textbf{94.20} & \cellcolor{bettergreen}\textbf{53.80} & \cellcolor{bettergreen}\textbf{40.10} & \cellcolor{bettergreen}\textbf{70.68} \\
    \hline
    \hline
    \multicolumn{8}{c}{\textit{\textbf{Qwen2.5-7B}}} \\
    \hline
    \multirow{2}{*}{\textbf{Base LLM}} 
    & \cellcolor{graycolor}Zero-shot &\cellcolor{graycolor} 84.20 & \cellcolor{graycolor}72.60 &\cellcolor{graycolor} 90.80 & \cellcolor{graycolor}61.20 & \cellcolor{graycolor}53.40 &\cellcolor{graycolor} 72.44 \\
    & Few-shot & 90.90 & 90.10 & 97.60 & 70.40 & 62.90 & 82.38 \\
    \cline{1-8}
    \multirow{2}{*}{\textbf{LLaMA}} 
    & \cellcolor{graycolor}CoT-SFT & \cellcolor{bettergreen}86.90 & \cellcolor{bettergreen}86.70 & \cellcolor{bettergreen}93.00 & \cellcolor{bettergreen}61.60 & \cellcolor{bettergreen}57.00 & \cellcolor{bettergreen}77.04 \\
    & FSLR & \cellcolor{bettergreen}\textbf{90.50} & \cellcolor{bettergreen}\textbf{91.10} & \cellcolor{bettergreen}\textbf{94.80} & \cellcolor{bettergreen}\textbf{67.20} & \cellcolor{bettergreen}\textbf{57.20} & \cellcolor{bettergreen}\textbf{80.16} \\
    \cline{1-8}
    \multirow{2}{*}{\textbf{Qwen}} 
    & \cellcolor{graycolor}CoT-SFT & \cellcolor{bettergreen}81.40 & \cellcolor{bettergreen}80.40 & \cellcolor{bettergreen}85.80 & \cellcolor{bettergreen}48.60 & \cellcolor{worsecolor}\textbf{58.80} & \cellcolor{bettergreen}71.00 \\
    & FSLR & \cellcolor{bettergreen}\textbf{86.50} & \cellcolor{bettergreen}\textbf{80.70} & \cellcolor{bettergreen}\textbf{94.00} & \cellcolor{bettergreen}\textbf{57.80} & \cellcolor{worsecolor}57.50 & \cellcolor{bettergreen}\textbf{75.30} \\
    \cline{1-8}
    \multirow{2}{*}{\textbf{Self}} 
    & \cellcolor{graycolor}CoT-SFT & \cellcolor{bettergreen}85.50 & \cellcolor{bettergreen}84.60 & \cellcolor{bettergreen}88.50 & \cellcolor{bettergreen}48.90 & \cellcolor{bettergreen}61.70 & \cellcolor{bettergreen}73.84 \\
    & FSLR & \cellcolor{bettergreen}\textbf{86.00} & \cellcolor{bettergreen}\textbf{90.80} & \cellcolor{bettergreen}\textbf{92.60} & \cellcolor{bettergreen}\textbf{54.70} & \cellcolor{bettergreen}\textbf{66.20} & \cellcolor{bettergreen}\textbf{78.06} \\
    \hline
    \hline
    \multicolumn{8}{c}{\textit{\textbf{Qwen3-4B}}} \\
    \hline
    \multirow{2}{*}{\textbf{Base LLM}} 
    & \cellcolor{graycolor}Zero-shot & \cellcolor{graycolor}78.10 & \cellcolor{graycolor}84.70 &\cellcolor{graycolor} 88.10 & \cellcolor{graycolor}66.30 & \cellcolor{graycolor}62.40 & \cellcolor{graycolor}75.92 \\
    & Few-shot & 88.40 & 84.80 & 95.90 & 70.30 & 56.00 & 79.08 \\
    \cline{1-8}
    \multirow{2}{*}{\textbf{LLaMA}} 
    & \cellcolor{graycolor}CoT-SFT & \cellcolor{bettergreen}81.10 & \cellcolor{bettergreen}85.70 & \cellcolor{bettergreen}89.30 & \cellcolor{bettergreen}62.10 & \cellcolor{bettergreen}58.50 & \cellcolor{bettergreen}75.34 \\
    & FSLR & \cellcolor{bettergreen}\textbf{89.90} & \cellcolor{bettergreen}\textbf{92.20} & \cellcolor{bettergreen}\textbf{96.00} & \cellcolor{bettergreen}\textbf{66.70} & \cellcolor{bettergreen}\textbf{66.00} & \cellcolor{bettergreen}\textbf{82.16} \\
    \cline{1-8}
    \multirow{2}{*}{\textbf{Qwen}} 
    & \cellcolor{graycolor}CoT-SFT & \cellcolor{bettergreen}89.90 & \cellcolor{bettergreen}87.70 & \cellcolor{bettergreen}94.30 & \cellcolor{worsecolor}\textbf{66.60} & \cellcolor{bettergreen}62.20 & \cellcolor{bettergreen}80.14 \\
    & FSLR & \cellcolor{bettergreen}\textbf{90.20} & \cellcolor{bettergreen}\textbf{90.00} & \cellcolor{bettergreen}\textbf{96.20} & \cellcolor{worsecolor}65.00 & \cellcolor{bettergreen}\textbf{63.70} & \cellcolor{bettergreen}\textbf{81.02} \\
    \cline{1-8}
    \multirow{2}{*}{\textbf{Self}} 
    & \cellcolor{graycolor}CoT-SFT & \cellcolor{bettergreen}89.10 & \cellcolor{bettergreen}90.10 & \cellcolor{bettergreen}96.20 & \cellcolor{worsecolor}\textbf{71.50} & \cellcolor{bettergreen}55.00 & \cellcolor{bettergreen}80.38 \\
    & FSLR & \cellcolor{bettergreen}\textbf{90.40} & \cellcolor{bettergreen}\textbf{92.10} & \cellcolor{bettergreen}\textbf{96.80} & \cellcolor{worsecolor}70.20 & \cellcolor{bettergreen}\textbf{64.50} & \cellcolor{bettergreen}\textbf{82.80} \\
    \hline
  \end{tabular}
  \caption{Out-of-distribution evaluation on five diverse benchmarks. Models are trained exclusively on SVAMP and evaluated on AsDIv, GSM8K, MAWPS, TabMWP, and GSM-Hard under zero-shot setting. Math-specialized models (DeepSeek-Math-7B and Qwen2.5-Math-7B) are evaluated zero-shot as reference baselines. Best results per setting are in \textbf{bold}. \colorbox{bettergreen}{Green cells} indicate FSLR outperforms CoT-SFT. \colorbox{worsecolor}{Red cells} indicate FSLR underperforms CoT-SFT.}
  \label{tab:cross-domain-svamp}
\end{table*}

\section{Prompt for Error Analysis}
\label{sec:error-analysis-prompt}

\noindent \textbf{Prompt Design.} To categorize reasoning errors, we use GPT-4o~\cite{gpt4o} with the following prompt template:

\begin{quote}
\small\ttfamily
You are analyzing mathematical reasoning errors to identify failures in understanding logical relationships between variables.

\textbf{Definition:} Genuine logical reasoning requires understanding the logical relationships between variables in a problem, including:
\begin{enumerate}
    \item Variable dependency: How variables depend on each other
    \item Condition-solution mapping: How given conditions constrain the solution approach
    \item Relevant information filtering: Which information is relevant vs. irrelevant to the solution
    \item Logical step dependency: Each reasoning step logically follows from previous steps
    \item Operation-relationship alignment: Choosing operations based on variable relationships, not surface-level keywords
\end{enumerate}

\textbf{Problem:} [problem text]

\textbf{Ground Truth Answer:} [ground truth]

\textbf{Model's Predicted Answer:} [prediction]

\textbf{Model's Reasoning Process:} [reasoning]

\textbf{Task:} Categorize this error into ONE of the following categories:

1. STRUCTURAL\_FAILURE: The error stems from misunderstanding logical relationships between variables

2. COMPUTATIONAL: The logical relationships are understood correctly, but arithmetic/calculation is wrong

3. COMPREHENSION: Misreading the problem statement itself

\textbf{Response Format:}\\
Category: [STRUCTURAL\_FAILURE/COMPUTATIONAL/COMPREHENSION]\\
Explanation: [One sentence explaining why this category was chosen]
\end{quote}

\section{Error Analysis}
\label{error}

\noindent We present additional error analysis visualizations for LLaMA-3.1-8B and Qwen3-4B models to complement the Qwen2.5-7B analysis in Figure~\ref{fig:failure_portion}.

\begin{figure}[h]
  \includegraphics[width=\columnwidth]{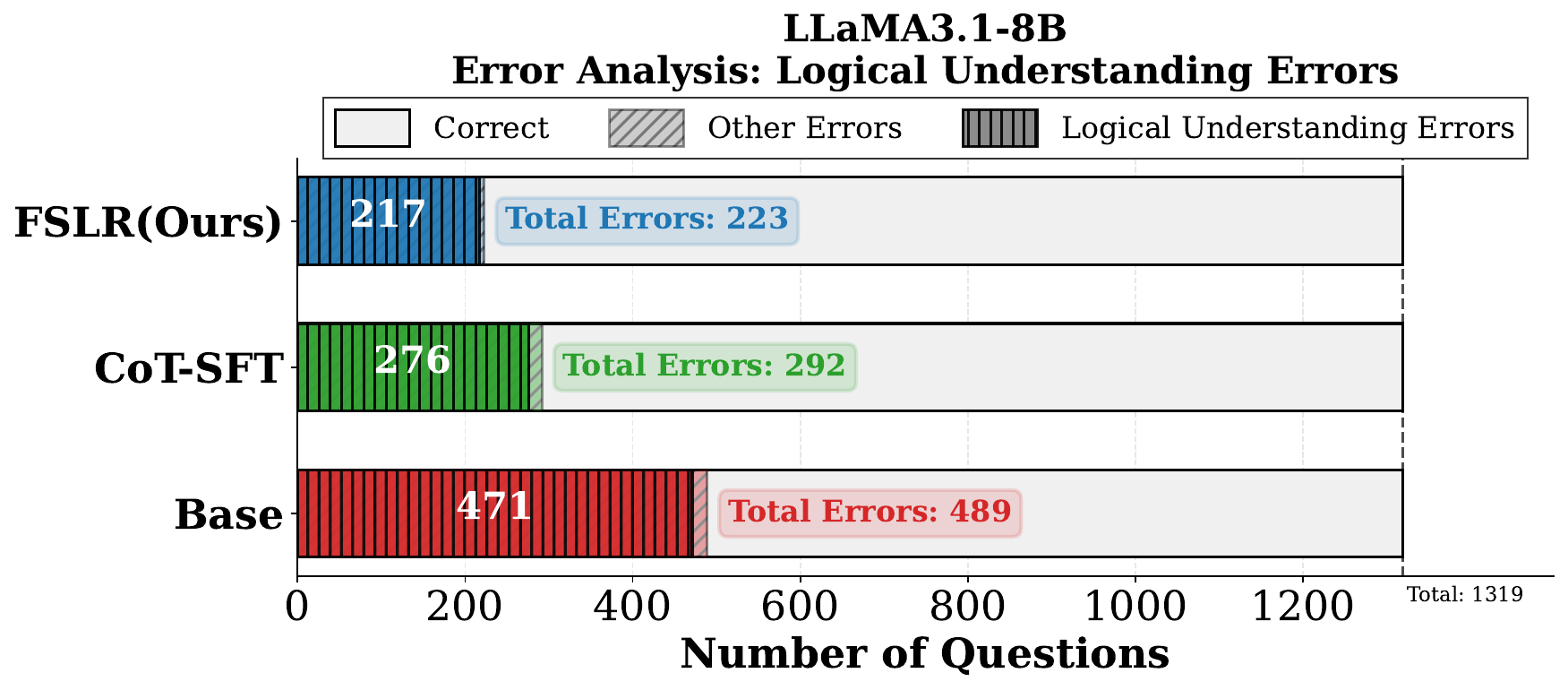}
  \caption{Error analysis on LLaMA3.1-8B comparing Base, CoT-SFT, and FSLR(Ours) models. Each bar shows the breakdown of correct predictions, logical relationship understanding errors, and other errors.}
  \label{fig:failure_portion_llama}
\end{figure}

\begin{figure}[h]
  \includegraphics[width=\columnwidth]{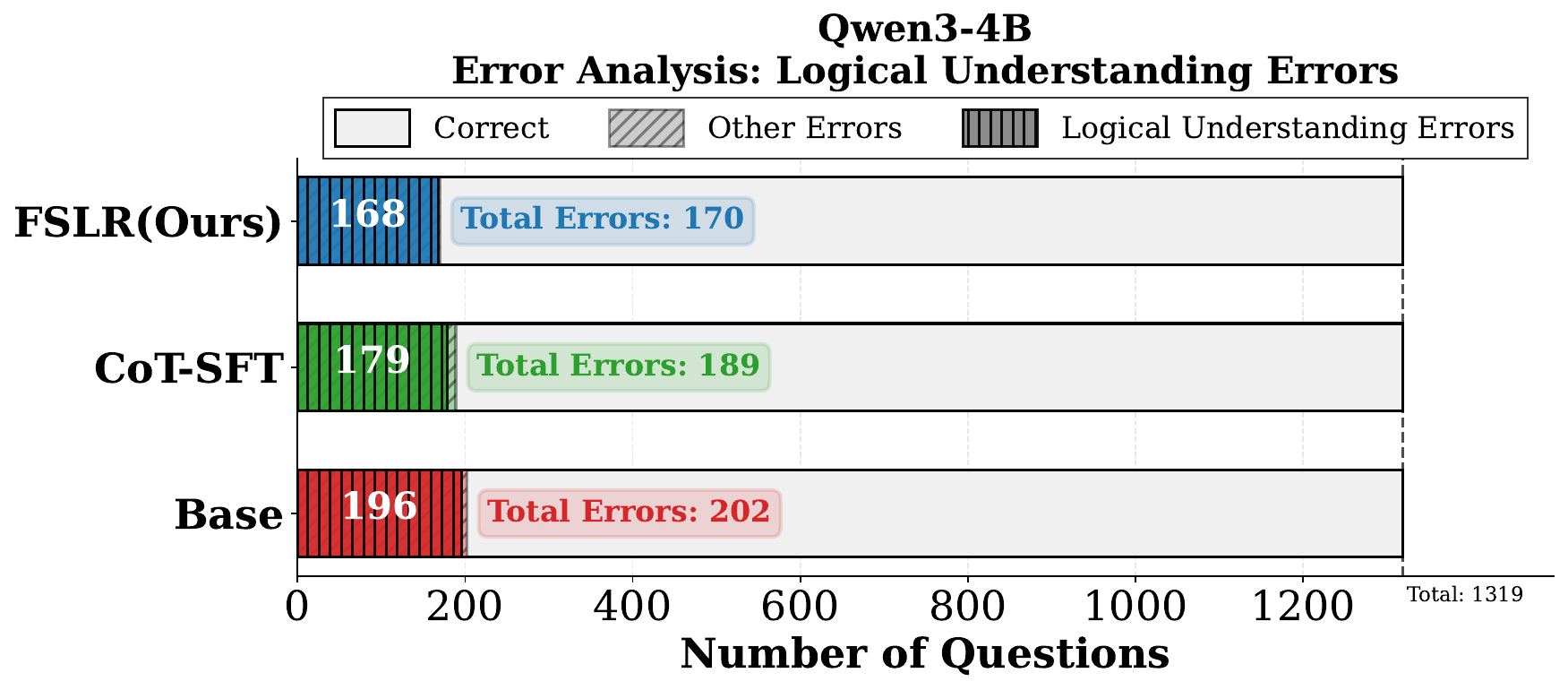}
  \caption{Error analysis on Qwen3-4B comparing Base, CoT-SFT, and FSLR(Ours) models. Each bar shows the breakdown of correct predictions, logical relationship understanding errors, and other errors.}
  \label{fig:failure_portion_qwen3}
\end{figure}

\noindent We observe consistent patterns across different models. On LLaMA-3.1-8B, FSLR reduces logical relationship understanding errors by 53.9\% compared to the base model and 21.4\% compared to CoT-SFT. On the stronger Qwen3-4B, logical relationship understanding errors still dominate (>90\% of failures), and FSLR continues to outperform CoT-SFT. These results confirm that regardless of base model strength, FSLR effectively targets the primary bottleneck in mathematical reasoning.

\section{Training Data Example}
\label{appendix:training-example}

To illustrate what the first-planning-step supervision looks like in practice, we provide a concrete example from our training data below.

\vspace{1em}
\noindent\textbf{Problem:} \textit{Betty picked 16 strawberries. Matthew picked 20 more strawberries than Betty and twice as many as Natalie. They used their strawberries to make jam. One jar of jam used 7 strawberries and they sold each jar at \$4. How much money were they able to make from the strawberries they picked?}

\vspace{0.5em}
\noindent\textbf{First-Planning-Step ($f_1$):} \textit{First, we need to calculate the number of strawberries Matthew picked, which is 16 (Betty's strawberries) + 20 (more than Betty) = ?}

\vspace{1em}

\noindent This example demonstrates how $f_1$ captures logical relationship understanding: the model identifies the relevant variables (Betty's 16 strawberries and the difference of 20), selects the appropriate operation (addition), and recognizes that ``20 more than'' implies an additive relationship. This isolation ensures the model is explicitly supervised on identifying the correct logical relationship from the problem statement.

\section{Comparison with Plan-and-Solve Fine-Tuning}
\label{plan_solve_baseline}

To investigate whether FSLR's performance gains stem specifically from the first-step design or from general planning supervision, we compare FSLR against a Plan-and-Solve fine-tuning baseline~\cite{plan_tuning}. Plan-and-Solve trains models to generate a complete plan before solving the problem, representing a broader form of planning supervision than FSLR's focused first-step approach.

\paragraph{Setup.} Using the planning generation prompt template from the original Plan-and-Solve work, we generate training data with LLaMA-3.1-70B-Instruct as the teacher model, maintaining the same experimental setup as FSLR. Results are reported on in-distribution benchmarks GSM8K and SVAMP.

\paragraph{Results.} As shown in Table~\ref{tab:plan-and-solve}, FSLR outperforms Plan-and-Solve by +1.20\% on average across all models and datasets. While Plan-and-Solve already improves over CoT-SFT (+3.60\% on average), FSLR achieves further gains, confirming that the first-step design contributes beyond general planning supervision. We attribute this to FSLR's more focused training signal: by isolating only the initial logical reasoning decision rather than generating a full plan, FSLR provides more direct supervision for logical relationship understanding without introducing the additional complexity of multi-step plan generation.

\begin{table*}[h]
  \centering
  \small
  \setlength{\tabcolsep}{4pt}
  \resizebox{0.95\textwidth}{!}{%
  \begin{tabular}{l|l|l|c|cc}
    \hline
    \textbf{Judge} & \textbf{Model} & \textbf{Method} & 
    \textbf{Total Errors} & \textbf{Logical Errors} & \textbf{Others} \\
    \hline
    \hline
    \multirow{9}{*}{\textbf{LLaMA-3.1-70B}} 
    & \multirow{3}{*}{LLaMA3.1-8B} 
        & \cellcolor{graycolor}Base    & \cellcolor{graycolor}489 & \cellcolor{graycolor}447 & \cellcolor{graycolor}42 \\
    &   & CoT-SFT                      & 292 & 282 & 10 \\
    &   & \cellcolor{graycolor}FSLR    & \cellcolor{graycolor}223 & \cellcolor{graycolor}215 & \cellcolor{graycolor}8  \\
    \cline{2-6}
    & \multirow{3}{*}{Qwen2.5-7B}  
        & Base    & 362 & 351 & 11 \\
    &   & \cellcolor{graycolor} CoT-SFT                      & \cellcolor{graycolor}296 & \cellcolor{graycolor}276 & \cellcolor{graycolor}20 \\
    &   & FSLR    & 197 & 185 & 12 \\
    \cline{2-6}
    & \multirow{3}{*}{Qwen3-4B}    
        & \cellcolor{graycolor}Base    & \cellcolor{graycolor}202 & \cellcolor{graycolor}187 & \cellcolor{graycolor}15 \\
    &   & CoT-SFT                      & 189 & 172 & 17 \\
    &   & \cellcolor{graycolor}FSLR    & \cellcolor{graycolor}170 & \cellcolor{graycolor}162 & \cellcolor{graycolor}8  \\
    \hline
    \hline
    \multirow{9}{*}{\textbf{Qwen2.5-72B}}   
    & \multirow{3}{*}{LLaMA3.1-8B} 
        & Base    & 489 & 443 & 46 \\
    &   & \cellcolor{graycolor}CoT-SFT                      & \cellcolor{graycolor}292 &\cellcolor{graycolor} 243 & \cellcolor{graycolor}49 \\
    &   & FSLR    & 223 & 206 & 17 \\
    \cline{2-6}
    & \multirow{3}{*}{Qwen2.5-7B}  
        & \cellcolor{graycolor}Base    & \cellcolor{graycolor}362 & \cellcolor{graycolor}330 & \cellcolor{graycolor}32 \\
    &   & CoT-SFT                      & 296 & 226 & 70 \\
    &   & \cellcolor{graycolor}FSLR    & \cellcolor{graycolor}197 & \cellcolor{graycolor}168 & \cellcolor{graycolor}29 \\
    \cline{2-6}
    & \multirow{3}{*}{Qwen3-4B}    
        & Base    & 202 & 164 & 38 \\
    &   & \cellcolor{graycolor}CoT-SFT                      & \cellcolor{graycolor}189 & \cellcolor{graycolor}162 & \cellcolor{graycolor}27 \\
    &   &FSLR    & 170 & 144 & 26 \\
    \hline
  \end{tabular}
  }
  \caption{Error attribution results using LLaMA-3.1-70B-Instruct and Qwen2.5-72B-Instruct as independent judges. Logical relationship understanding errors consistently account for over 90\% of failures across all judges, models, and methods, consistent with the GPT-4o-based analysis in the main paper.}
  \label{tab:judge}
\end{table*}

\section{Pass@k Evaluation on GSM8K}
\label{pass_k}
To investigate whether FSLR's improvements genuinely expanding model capability boundarie, we evaluate using Pass@$k$ metrics on GSM8K, which measure whether the correct answer appears in $k$ attempts and thus reflect the upper bound of model capability. As shown in Table~\ref{tab:passk}, FSLR consistently outperforms both Base and CoT-SFT across all $k$ values. The gains persist even as $k$ increases (+1.74\% at Pass@4, +1.16\% at Pass@8, +0.68\% at Pass@16), demonstrating that FSLR expands the model's capability boundary.

\begin{table}[h]
  \centering
  \small
  \setlength{\tabcolsep}{4pt}
  \resizebox{0.95\columnwidth}{!}{%
  \begin{tabular}{l|ccc}
    \hline
    \small
    \textbf{Metric} & \textbf{Base} & \textbf{CoT-SFT} & \cellcolor{graycolor}\textbf{FSLR} \\
    \hline
    Pass@4  & 93.10 & 94.30 & \textbf{96.04} \\
    Pass@8  & 96.21 & 96.63 & \textbf{97.79} \\
    Pass@16 & 97.88 & 97.80 & \textbf{98.56} \\
    \hline
  \end{tabular}
  }
  \caption{Pass@$k$ evaluation on GSM8K for LLaMA3.1-8B with 
  LLaMA-3.1-70B-Instruct as teacher. FSLR consistently improves 
  over Base and CoT-SFT across all $k$ values.}
  \label{tab:passk}
\end{table}

\section{Out-of-Distribution Performance (SVAMP Training)}
\label{sec:ood-svamp}

Table~\ref{tab:cross-domain-svamp} presents out-of-distribution results where models are trained exclusively on SVAMP and evaluated on five diverse benchmarks: AsDIv, GSM8K, MAWPS, TabMWP, and GSM-Hard. This complementary experiment validates whether the benefits of FSLR training generalize when using a different, smaller training dataset.

\noindent\textbf{FSLR maintains strong generalization even with limited training data.} Despite SVAMP being a smaller dataset than GSM8K, FSLR consistently outperforms CoT-SFT across nearly all configurations. On LLaMA-3.1-8B, FSLR achieves substantial improvements: +10.2\% on AsDIv (85.9\% vs 75.7\% with LLaMA teacher), +16.1\% on MAWPS (94.4\% vs 78.1\% with Self teacher), and +8.6\% on TabMWP, averaging +8.92\% improvement across all OOD datasets with the LLaMA teacher. Qwen2.5-7B and Qwen3-4B show similar trends, achieving average gains of +4.22\% and +2.42\%, respectively. These results demonstrate that FSLR's effectiveness is not dependent on large-scale training data. Consistent with GSM8K training results, \textbf{FSLR-trained models surpass math-specialized models}. The results suggest that teaching models \textit{how to initiate reasoning} through logical relationship understanding is more effective for generalization than exposing them to large volumes of mathematical content. This reinforces our core hypothesis that the bottleneck in mathematical reasoning lies in the ability to identify and apply appropriate reasoning strategies from the problem's logical structure.

\section{Reliability of Error Attribution: Multi-Judge Analysis}
\label{multi_judge}

Table~\ref{tab:judge} presents error attribution results using LLaMA-3.1-70B-Instruct and Qwen2.5-72B-Instruct as independent judges, complementing the GPT-4o-based analysis in the main paper. Logical relationship understanding errors consistently account for over 90\% of failures across all judges, models, and methods, confirming the robustness of our error attribution.

\section{Complete Case Study}
\label{sec:case-study}
To illustrate how FSLR improves reasoning through better understanding of logical relationships, we present one example from GSM8K where LLaMA-3.1-8B-Instruct model and CoT-SFT both fail, while FSLR succeeds.

\noindent\textbf{Problem:} \textit{Shiela bought 5 cell phones for \$150 each for a 3-month installment. A 2\% interest will be charged for each unit. How much will Shiela pay each month for 3 months?}

\noindent\textbf{Ground Truth:} \textit{A \$150 × 2/100 = \$3 interest will be charged for each unit. So, each cell phone costs \$150 + \$3 = \$153. The total amount to be paid for 5 cell phones is \$153 × 5 = \$765. Thus, Shiela will pay \$765/3 = \$255 each month. \textbf{Answer: \$255}}

\noindent\textbf{LLaMA-3.1-8B-Instruct Response:} \textit{Step 1: Calculate the total cost of the cell phones. Total cost = 5 × \$150 = \$750. Step 2: Calculate the interest charged for each unit. Interest = 2\% of \$150 = 0.02 × \$150 = \$3. Step 3: Calculate the total interest charged for 3 months. Total interest = 3 × \$3 = \$9. Step 4: Calculate the total amount Shiela will pay each month for 3 months. Total amount = Total cost + Total interest = \$750 + \$9 = \$759. Step 5: Calculate the monthly payment. Monthly payment = Total amount / 3 = \$759 / 3 = \$253. \textbf{Answer: \$253} }

\noindent\textbf{CoT-SFT Response:} \textit{First, we need to calculate the total cost of the cell phones. Shiela bought 5 cell phones for \$150 each, so the total cost is: 5 × \$150 = \$750. Next, we need to calculate the interest charged for each unit. The interest rate is 2\% per unit, and there are 5 units (cell phones). So, the total interest charged is: 5 × 2\% = 10\%. Now, we need to calculate the total amount Shiela needs to pay, including the interest. To do this, we'll add the total cost of the cell phones to the total interest charged: \$750 (total cost) + \$750 × 10\% (interest) = \$750 + \$75 = \$825. Since Shiela is paying in installments over 3 months, we need to divide the total amount by 3 to find out how much she'll pay each month: \$825 ÷ 3 = \$275. \textbf{Answer: \$275}}

\noindent\textbf{FSLR Response:} \textit{First, calculate the total cost of the cell phones: 5 cell phones × \$150 per cell phone = \$750. Then, calculate the total interest: 2\% of \$750 = 0.02 × \$750 = \$15. Now, calculate the total amount Shiela will pay: \$750 (cost of cell phones) + \$15 (interest) = \$765. Finally, calculate the monthly payment: \$765 / 3 months = \$255 per month. \textbf{Answer: \$255}}\\

\end{document}